\documentclass[12pt]{article}
\usepackage[utf8]{inputenc}

\usepackage{titling}
\usepackage{blindtext}
\usepackage{authblk}
\usepackage{setspace}
\usepackage[letterpaper, margin=1.0in]{geometry}
\usepackage{graphicx}
\usepackage{amsmath}
\usepackage{amssymb} % For checkmark
\usepackage{tikz}
\usepackage{enumerate}

\usepackage{url}
\usepackage{setspace}
\usepackage[round,sort,comma,authoryear]{natbib}
\usepackage{authblk}
\usepackage{blindtext}

\newcommand{\ie}{\textit{i}.\textit{e}.}
\newcommand{\eg}{\textit{e}.\textit{g}.}

\begin{document}

\title{Image as Data: Automated Visual Content Analysis \\for Political Science}
%\title{Images as Data: Automated Visual Data Analysis for Political Science}
%\title{Images as Data: Computer Vision and Deep Learning for Political Science}
%\title{Automated Visual Data Analysis for Political Science \\ with Deep Learning}
%\title{Computer Vision Using Deep Learning: Automated Visual Analysis for Political Science}

%\author{Anonymous Submission}
%\author{Jungseock Joo\thanks{jjoo@comm.ucla.edu} }
%\author{Zachary C. Steinert-Threlkeld\thanks{zst@luskin.ucla.edu} }
%\affil{University of California, Los Angeles}

\author[1]{Jungseock Joo}
\author[2]{Zachary C. Steinert-Threlkeld}
\affil[1]{Department of Communication}
\affil[2]{Department of Public Policy}
\affil[ ]{University of California, Los Angeles}
\affil[ ]{{jjoo@comm.ucla.edu, zst@luskin.ucla.edu}}

\date{}

%\date{\today}
\maketitle
\begin{abstract}
% AJPS abstract 150 words limit
Image data provide unique information about political events, actors, and their interactions which are difficult to measure from or not available in text data. This article introduces a new class of automated methods based on computer vision and deep learning which can automatically analyze visual content data. Scholars have already recognized the importance of visual data and a variety of large visual datasets have become available. The lack of scalable analytic methods, however, has prevented from incorporating large scale image data in political analysis. This article aims to offer an in-depth overview of automated methods for visual content analysis and explains their usages and implementations.  We further elaborate on how these methods and results can be validated and interpreted.  We then discuss how these methods can contribute to the study of political communication, identity and politics, development, and conflict, by enabling a new set of research questions at scale. 

%Image data provide unique information about political events, actors, and their interactions which are difficult to measure from or not available in text data. This article introduces a new class of automated methods based on computer vision and deep learning which can automatically analyze the visual content of image data. Scholars have already noted the importance of visual data and the recent surge in social media especially has made an enormous amount of visual data available for research. Due to the lack of a scalable analytic method, however, it has been impossible to incorporate large scale image data in political analysis. This article therefore aims to offer an in-depth overview of automated methods for visual content analysis and explains their usages and implementations, focusing on deep learning and convolutional neural networks.  We further elaborate on how these methods and their results can be validated and interpreted.  We then discuss how these methods can contribute to the study of political communication, identity and politics, development, and conflict.
%, and we develop an extended example of how they can generate protest event data.  
%These methods make it possible to ask a new set of research questions at scale, just as automated text content analysis has contributed to the field in the past decade.  
\end{abstract}

%Word count: 7447 (text) + 344 (caption and footnotes) + 2189 (references) = 9980

\clearpage
\section{Introduction: From Text to Image}

In 1976, photographs of President Gerald Ford failing to husk a tamale may have cost him the presidential election.  In 1988, Democratic front runner Gary Hart was felled by a photo of him with a mistress; the man who became the nominee, Michael Dukakis, by an awkward photo riding an M1 Abrams tank.  In 2004, candidate John Kerry was photographed wind surfing, cementing his reputation as an effete elite. In 2010, video of a self-immolated fruit vendor spread throughout Tunisia, sparking the Arab Spring.  Visual communication is a powerful component of politics, and new methods from computer vision and deep learning are enabling political scientists to better understand its power. 

Political scientists have developed and applied advanced computational techniques to large scale text corpora, such as party manifestos \citep{Laver2003a,Mikhaylov2011}, Congressional press releases \citep{grimmer2010bayesian,Grimmer2013}, news articles \citep{Hopkins2010}, and survey responses \citep{Hobbs2017a}.  These methods have advanced in response to the growing availability of textual data in a quantity that overwhelms manual analysis.  Automated content analysis methods, therefore, have been widely used by political scientists in the past decade, ranging from simple keyword based methods to topic modeling or sentiment analysis and opinion mining. 

This article argues that we are at a similar juncture with visual data.  While political scientists have long understood the importance of imagery in politics, analysis has consisted of the manual collection and annotation of data.  The labor intensiveness of the collection process has limited the external validity of studies and prevented answering certain kinds of questions.  Moreover, the rise of the internet, and social media in particular, provides political scientists with vastly more visual data.  There is now more data than a single team can analyze manually, requiring the adoption of new methodologies.  This paper introduces political scientists to these methodologies.

Visual data are characterized by several key features, distinct from text data, as summarized in Table~\ref{tab:textimage}. Due to these features, it is often challenging or impossible to apply the same machine learning techniques for text data to visual data. The most critical distinction between them is that an image is a two dimensional array of pixels and each pixel carries no semantic meaning, as opposed to text data whose atomic elements are words. A single word can provide a great deal of semantic information, \eg, ``Trump'' or ``election,'' and a simple string comparison operation allows to access the information. In contrast, in visual analysis one has to process a huge number of meaningless pixels to detect and identify people, objects and events just to be on par with the starting point of text analysis. Recognizing elementary content, \ie, visual ``words,'' from an image is, however, extremely difficult. This technical difficulty has been the main obstacle to research inquires involving quantitative analysis of visual data on a large scale.

%The information present in visual data has not been explored mainly due to the technical difficulty of analyzing them compared to verbal, text data. 

\begin{table}
\caption {Distinct Characteristics between Text and Image Data} 
\label{tab:textimage} 
\begin{center}
    \begin{tabular}{| p{0.45\textwidth} | p{0.45\textwidth} |}
	\hline
	\multicolumn{1}{|c|}{Text} & \multicolumn{1}{|c|}{Image} \\
	\hline
    \begin{itemize}
  		\item One dimensional: \newline a sequence of words
  		\item Low uncertainty at word level
        \item Small size; easy to transfer and store
        \item Known dictionary
        \item Language specific
        \item Elaborative 
        \item More logical
	\end{itemize}
    &
    \begin{itemize}
  		\item Two dimensional: \newline an array of pixels
  		\item High uncertainty at any level
        \item Bigger size
        \item Unknown dictionary
        \item Universal
        \item Intuitive and immediate
        \item More emotional
	\end{itemize} \\
    \hline
    \end{tabular}
\end{center}
\end{table}

This paper introduces recent breakthroughs in computer vision and machine learning and demonstrates how they can be applied to political science research.  The new approach, colloquially known as deep learning, represents significant improvements in learning from big data, with the help of increased hardware capabilities, especially the prevalence of graphical processing units (GPUs).  
%The falling cost of computing coupled with massive datasets means these techniques will become mainstream shortly.  Our goal is to hasten that process. 
In addition to explaining how deep learning works, tasks for which it is well-suited, and training and validation, this paper suggests substantive research areas in which these techniques will prove useful.  
Finally, an analysis of protests in South Korea and Hong Kong using social media images is presented as a demonstration of this promising methodology. 
%Finally, an analysis of protests in South Korea and Hong Kong using social media images is presented in the Supporting Information as a demonstration of this promising methodology. 

%In addition to explaining how deep learning works, tasks for which it is well-suited, and training and validation, this paper suggests substantive research areas in which these techniques will prove useful.  Images containing faces, buildings, or vehicles can provide demographic and socioeconomic data, useful for scholars studying how outcomes such as legislator responsiveness varies.  These data could be particularly useful for inquiries concerning development or civil conflict, as they may provide finer temporal and geographic resolution concerning economic conditions than government statistics or nighttime light data.  These techniques can also be applied to newspaper images, television broadcasts, and images shared on websites, permitting political communication scholars to study how coverage varies by topic, medium, and company as well as what types of images increase or decrease support for an issue.  Finally, deep learning holds promise for the creation of protest event data, as it can generate estimates of crowd size and violence, measurements that are much harder using text corporate.  This promise is demonstrated with an analysis of protests in South Korea, Hong Kong, Venezuela, and Egypt.

\section{Computer Vision and Deep Learning}
\subsection{Goals}
Computer vision is an interdisciplinary branch of study crossing computer science, statistics, cognitive science and psychology. The primary goal of computer vision is automated understanding of visual content, \ie, to replicate human vision abilities such as face recognition or object detection by computational models.
%which should obtain those abilities through learning. 

The human vision system is versatile, complicated, and not fully understood, and the computer vision systems cannot simply reconstruct the mechanisms of human vision. Therefore, the literature has mostly focused on using statistical inference and machine learning approaches to deal with noisy inputs and discover meaningful patterns. In practice, this pipeline usually consists of collecting a large amount of visual data, manually labeling the data, and training a model (estimating its learnable parameters) which can best explain the observed data.

The insufficient reliability and accuracy of computer vision based methods was the primary limiting factor to practical applications -- including political analysis of visual content -- until the field made a dramatic leap forward with the advances in deep learning based approaches, which will be elaborated in the following sections.

%The following subsections explains these advances.  Section \ref{sec:tasks} highlights how the current approach is applied broadly, Section \ref{sec:trainValidate} explains training and validating deep learning models, Section \ref{sec:appPolisci} discusses how the current approach can contribute to political science, and Section \ref{sec:appEventsData} demonstrates how computer vision can generate event data.

\subsection{Deep Learning and Hierarchical Representations}
Deep learning refers to a class of machine learning methods which utilizes hierarchical, multi-layered models.\footnote{Computer vision tries to solve visual \textit{problems} with any kinds of methods and deep learning offers efficient \textit{methods} to any kinds of data. }
%\footnote{The distinction between computer vision and deep learning is that computer vision tries to solve visual problems with any kinds of automated methods and deep learning offers a specific set of machine learning methods to any kinds of data (visual, textual, audio, numeric, etc). }
In contrast to shallow or flat models such as linear regression in which output variables can be directly computed from input variables, ``deep'' models employ repetitive structures with multiple layers\footnote{A layer means a separate operation in a network. It will be further elaborated shortly.} such that the final outputs of the model are obtained through a sequence of operations applied to the input data and intermediate results. 

%the output of any intermediate layer becomes the input for the next layer and

In machine learning, hierarchical model structures are commonly used, \eg, topic models such as LDA  \citep{blei2003latent} or PLSA \citep{hofmann1999probabilistic}. These models incorporate different levels of representations which capture structured and global information (\eg, topic) as well as local information (\eg, words) from input data. 

%is that higher level representations, hyperparameters, govern more structured and global information whereas lower level representations or features deal with local and specific information from the input data. 

Deep learning-based methods profit from the same hierarchical structure, but they employ a larger number of consecutive layers.  These extra layers add the ``deep'' to the learning. Indeed, the success of deep learning is related to the depth of the models, as additional layers can achieve a higher expressiveness and capture more complex data distributions than what shallower models can  \citep{delalleau2011shallow,eldan2016power,poggio2017and}. 
%\citep{delalleau2011shallow,eldan2016power,raghu2017expressive,mhaskar2017and}. 
%Deeper models permit models to better fit the data on which they train.
Furthermore, such complex internal structures and interactions are directly learned from data rather than manually defined. This is also advantageous when applied to complex data such as images.

\subsection {Artificial Neural Network}
While there have been different models proposed in the deep learning literature, artificial deep neural networks (DNN) are the most popular branch of deep models and have been used in a number of areas including computer vision, audio processing, natural language processing, robotics, bioengineering, and medicine. This subsection describes a general neural network, and Section \ref{sec:CNN} discusses its variant, a convolutional neural network (CNN).  The convolutional neural network is commonly used in computer vision applications with two-dimensional inputs.

Artificial neural networks represent complex concepts, like the probability an image contains a human face, as a system of connections between elementary \textit{nodes}; the collection of nodes and connections is the neural network. Each node, also called a \textit{neuron} or an \textit{unit}, in this system only performs simple computations and interacts with a few other nodes. Nevertheless, the network of a large number of nodes enables complex data modeling through their interactions. 

\begin{figure}[h]
\centering
\includegraphics[width=0.8\textwidth]{{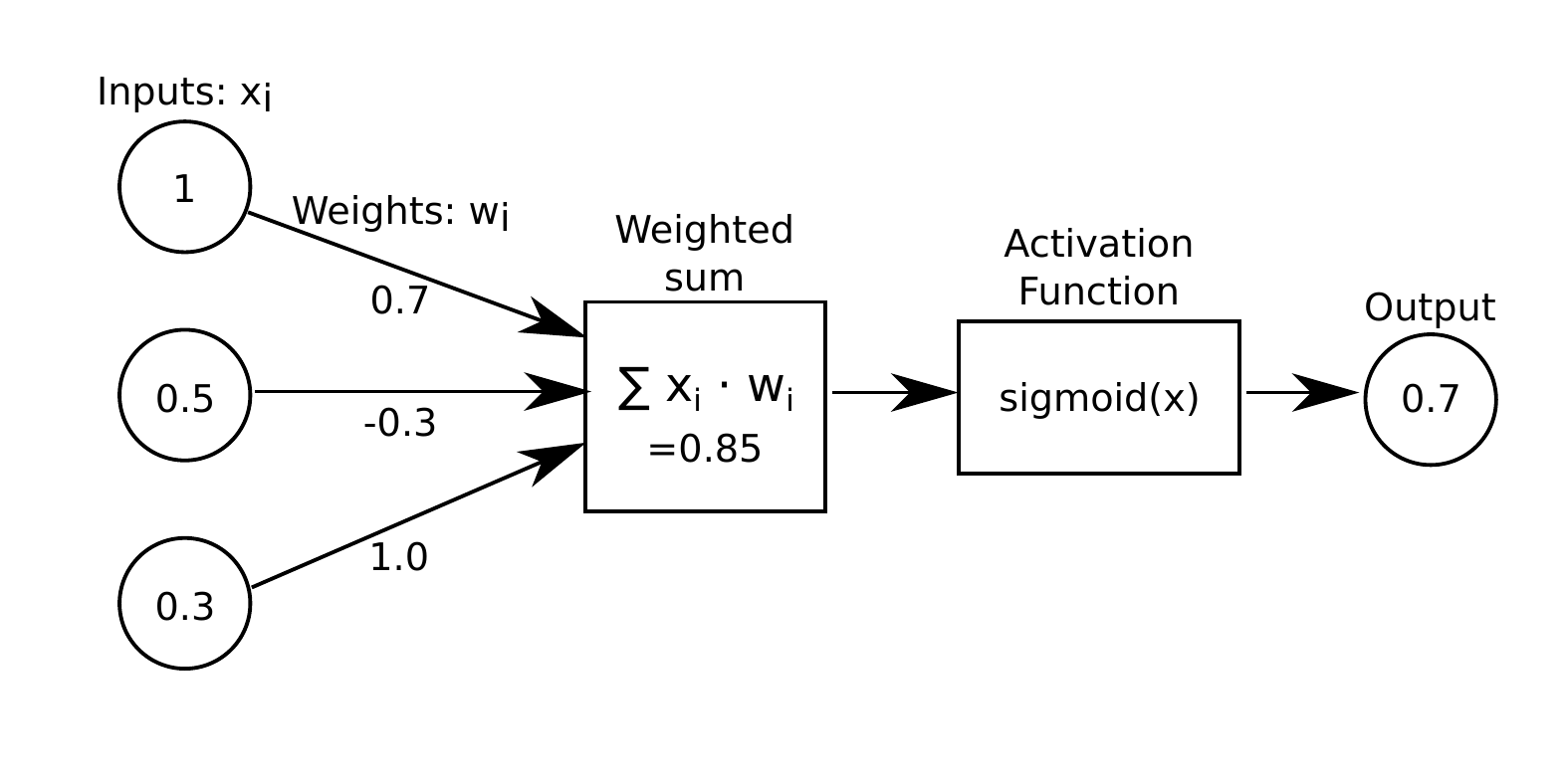}}
\caption{An example computation in a node and its connected nodes.  }
\label{fig:neuron}
\end{figure}

Figure~\ref{fig:neuron} shows an example configuration of a node and its connected nodes. Each node takes input values from its input nodes and evaluates a weighted sum using weights associated with edges (in this example, $1 \cdot 0.7 + 0.5 \cdot -0.3 + 0.3 \cdot 1.0 = 0.85$). Typically this value is transformed by a non-linear activation function, \eg, sigmoid, and then passed to output node.

\begin{figure}[h]
\centering
\includegraphics[width=0.7\textwidth]{{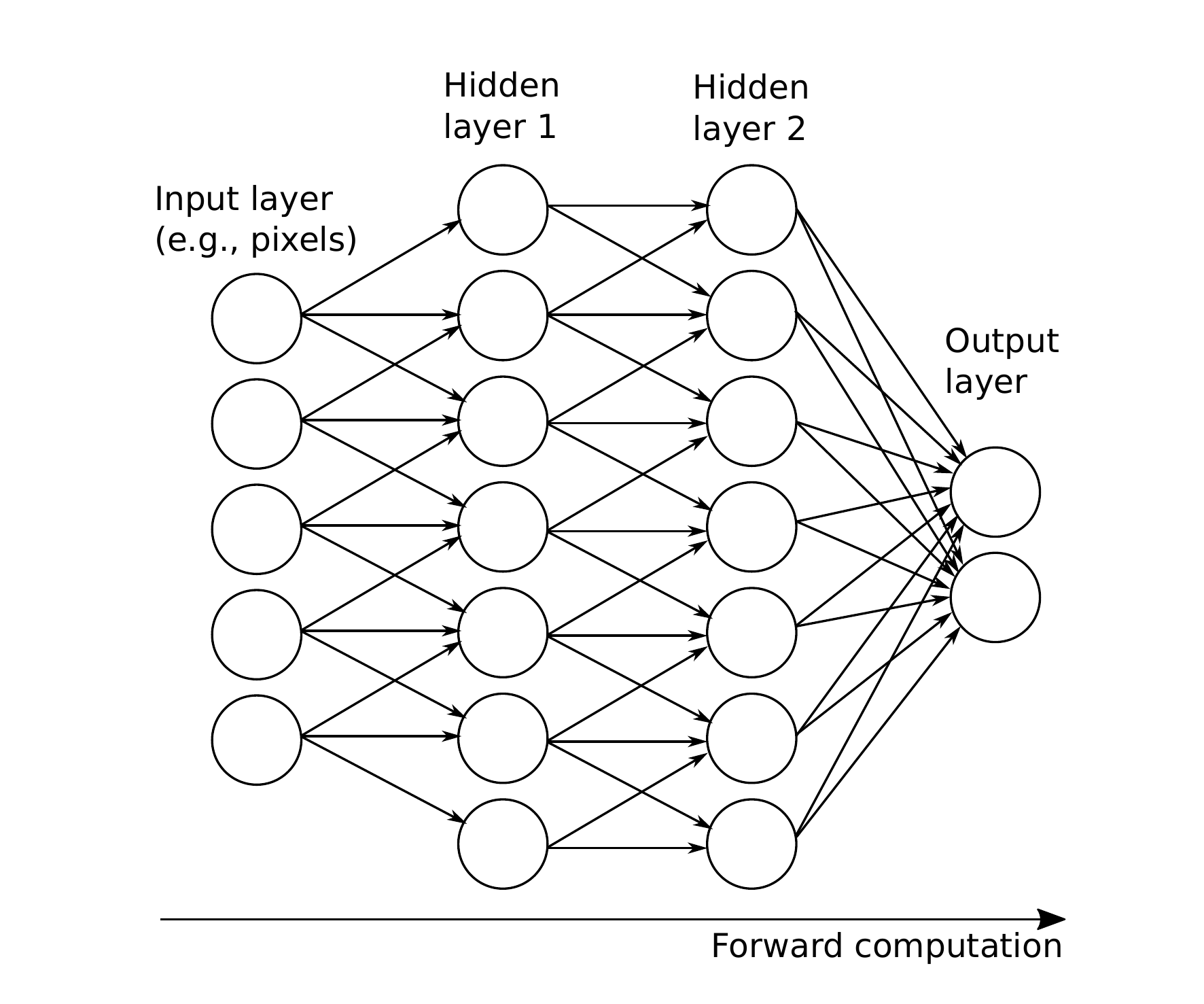}}
\caption{An example architecture of a neural network with an input layer, an output layer, and two hidden layers.  
}
\label{fig:layers}
\end{figure}

Figure~\ref{fig:layers} shows an example architecture of a neural network with several layers. Neural networks with multiple hidden layers are considered ``deep.'' A \textit{layer} in a neural network is a set of nodes which takes inputs from the nodes in the previous layer and deliver outputs to the nodes in the next layer.  When a network is visualized as Figure~\ref{fig:layers}, a column of nodes is a layer, and the number of columns is the number of layers.  Inputs to the whole network therefore undergo several steps of transformations through layers until they reach the output layer of the network.  The output layer is the network's final layer, and it contains one node per desired label in case of classification.

Hidden layers are intermediate layers between the input and output layers in a network whose true values are not observed during training. They play a critical role in modeling complex concepts by giving an expressive power to deeper networks. Studies have shown, both experimentally and theoretically, that the more layers a neural network has, the better performance it can achieve \citep{eldan2016power, poggio2017and}. A drawback of having too many layers is that it is more difficult to train such a model, \ie, vanishing gradients \citep{bengio1994learning}\footnote{Networks are trained by a gradient descent method with backpropagation, and the gradients become smaller as it goes back through more layers, which makes it very hard to update the parameters. }.

Training a neural network is a task to find optimal values for edge weights (i.e., in Figure~\ref{fig:neuron}). In most cases, objective functions of neural networks are non-convex and training is conducted by a gradient descent method with the backpropagation algorithm \citep{lecun1989backpropagation}, alternating between forward and backward passes. In the ``forward'' pass, given an input value, the network evaluates the output and computes the loss function based on the ground truth output value. In the ``backward'' pass, the gradient of the loss function is propagated backward by the chain rule and model weights are updated accordingly.

\subsection {Convolutional Neural Network}
\label{sec:CNN}
\begin{figure}[htpb!]
\centering
\includegraphics[width=0.9\textwidth]{{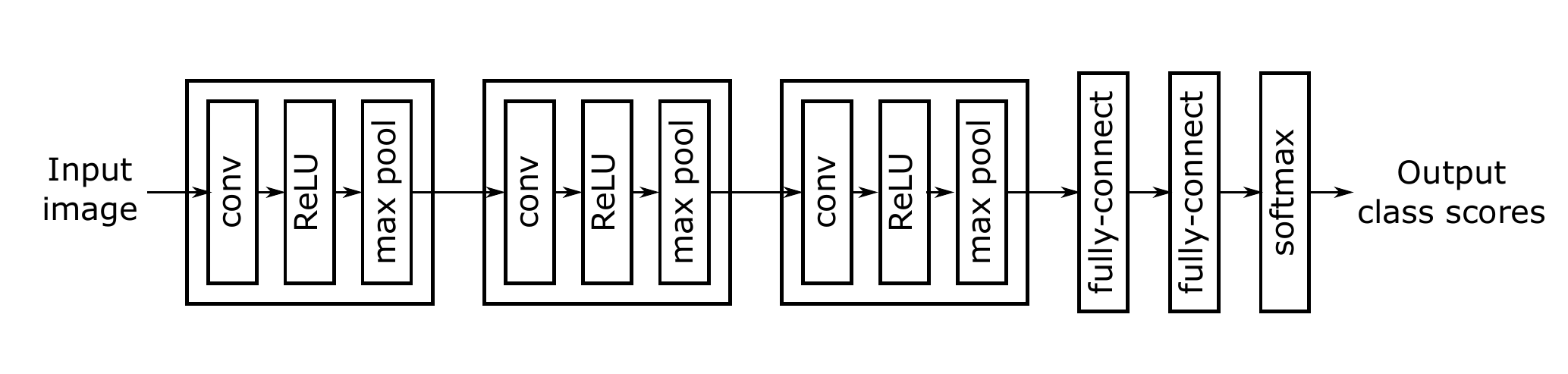}}
\caption{An example of a convolutional neural network architecture. 
}
\label{fig:cnn}
\end{figure}

Figure~\ref{fig:cnn} illustrates an example configuration of a typical convolutional neural network (CNN) for classification.  \cite{lecun1989backpropagation} first proposed the CNN structure with an efficient learning algorithm based on the backpropagation.  Since \cite{krizhevsky2012imagenet} showed an impressive performance, it has become the \emph{de facto} standard method for image classification. CNNs have a repetitive structure with several important layers: the convolutional layer, nonlinear layer (ReLU, in Figure \ref{fig:cnn}), pooling layer, and fully connected layer. This subsection describes each in turn.

% * <zst@luskin.ucla.edu> 2018-09-05T15:52:27.790Z:
% 
% What about briefly describing the softmax layer here?  Figure 3 shows a softmax layer, but we do not discuss it in its own subsection.  It feels like we forgot about it because the orther layers have their own subsections.  I see we discuss it later - "The softmax function is commonly" - so maybe that text can be moved?  Basically, I think Section 2.4 needs to explain the softmax layer.
% 
% ^.

\subsubsection{Convolutional layer}
\label{sec:convlayer}
A convolutional layer in CNNs performs a convolution operation to the input  to the layer, either raw image data or an output from the previous layer. 
Convolution is widely used in signal processing for transforming or comparing sequence data.  For example, one can reduce noise in a signal by convolving it with a Gaussian filter, which will smooth out the original signal by blending the original value at time $t$ with other values at adjacent time points around $t$.  Convolutino is also used for template matching, where one wants to detect the specific part of a signal that matches a pre-defined template: this is the main purpose of convolutional layers in CNNs. 

Formally, the convolution of two functions, $f$ and $g$, is another function defined by
\begin{equation}
(f \text{ * } g ) (t) 
= %\overset{\underset{\mathrm{def}}{}}{=} 
\int f(x) \cdot g(t-x) dx.
\end{equation}

The second function, $g(t)$, is called a \textit{kernel}. Note that the kernel is flipped ($g(t-x)$) by the definition of convolution. In a discrete case, this measures the sum of element-wise multiplication between two functions, with one function being shifted over time, such that
\begin{equation}
(f \text{ * } g ) (t) 
= %\overset{\underset{\mathrm{def}}{}}{=} 
\sum_{x} f(x) \cdot g(t-x).
\end{equation}

\begin{figure}[h]
\centering
\includegraphics[width=0.8\textwidth]{{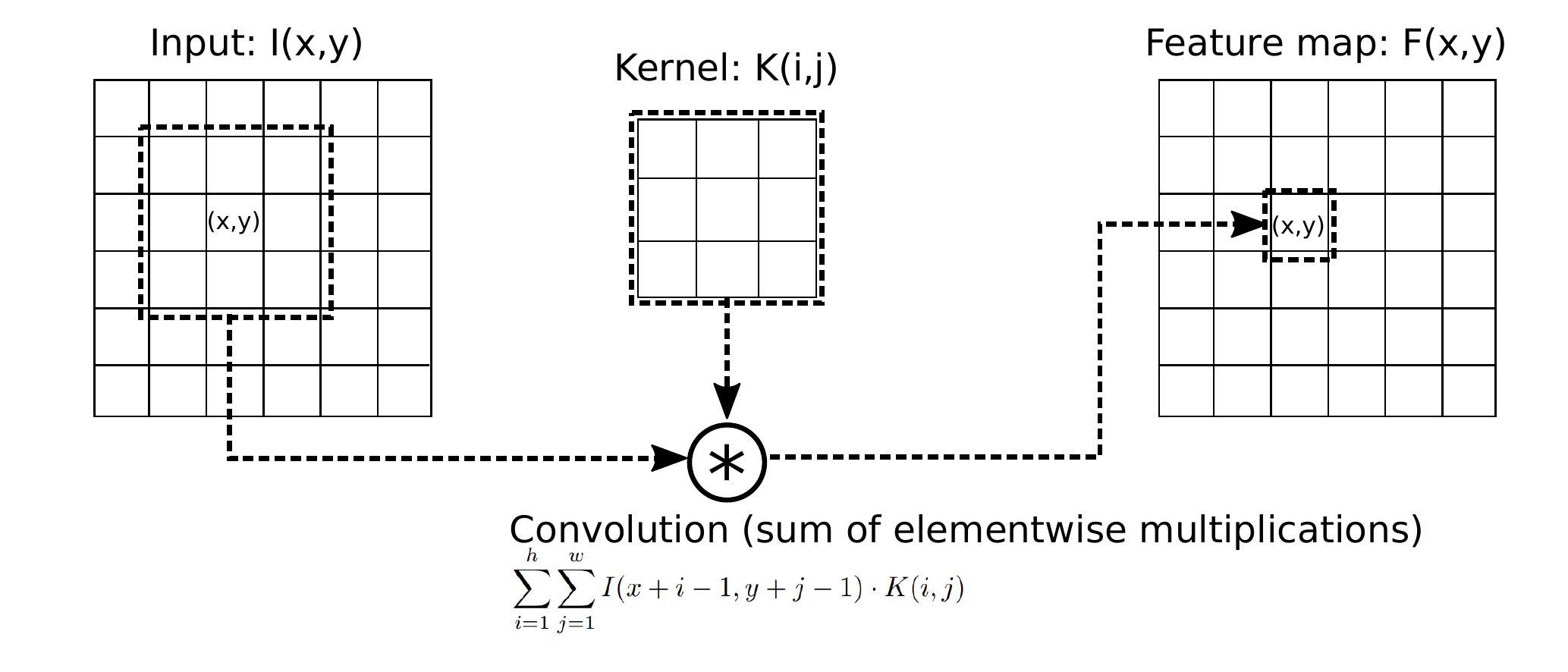}}
\caption{Illustration of computations in a convolutional layer. 
}
\label{fig:conv}
\end{figure}

Each convolutional layer in a CNN uses a convolution operation in order to compare the input data with the kernels that are stored in the model. The kernel is also called the \textit{filter} in the deep learning literature. % The input, e.g., an image, is usually multidimensional.  
In practice, the kernel is not flipped in computation in most implementations as it is unnecessary for the purpose of CNN\footnote{The parameters will be learned in the same way irrespective of the flipping direction.}. This will give us a slightly modified definition of convolution of a two-dimensional input $I$ and a two-dimensional kernel $K$ in CNN: 

\begin{equation}
F(x, y) = 
(I \text{ * } K ) (x,y) \overset{\underset{\mathrm{def}}{}}{=} 
\sum_{i=1}^{h}\sum_{j=1}^{w} I(x+i-1, y+j-1) \cdot K (i, j).
\end{equation}
$I(x,y)$ and $K(x,y)$ denote the element in $x$th row and $y$th column in the matrices $I$ and $K$. $h$ and $w$ denote the height and width of the kernel $K$, and, typically, CNNs use square kernels ($h=w$). The result of the convolution is another 2D array, $F$, which is called a \textit{feature map}. This is the output of the convolutional layer. This computation is performed on every location in an input map and the result is stored in the same location in the output feature map (See Figure~\ref{fig:conv}).  

Most images are three-dimensional data with two spatial dimensions and an additional dimension of color channel (e.g., RGB).  Also, intermediate feature maps in each layer are three-dimensional as each feature map (or each channel) corresponds to the response from a specific kernel (filter)\footnote{For example, a layer may contains 10 filters describing the shape of digits, 0-9. Then the resulting output (10 $\times$ W $\times$ H) will have 10 feature maps (or 10 channels) of the same size, each recording the response for one digit.}. The entire weight parameters of each convolutional layer ($K$) are therefore represented by a four-dimensional array of size ($w, h, m, n$), where $m$ is the number of channels of the input (the number of channels in the previous convolutional layer) and $n$ is the number of channels in the current layer. 
%Note that the convolution is still two-dimensional because it does not convolve in the channel dimension.
The feature map for each channel will therefore be obtained as follows: 
\begin{equation}
F(x, y,c') = 
\sum_{c=1}^{m}\sum_{i=1}^{h}\sum_{j=1}^{w} I(x+i-1, y+j-1, c) \cdot K (i, j, c, c').
\label{eq:conv4}
\end{equation}

%Convolutional neural networks have several important properties that we will briefly review as follows. 

Convolutional layers enable the following two key properties of convolutional neural networks.

\textbf{Weight sharing}. In Equation~\ref{eq:conv4}, the kernel is invariant to the location of each input node ($x, y$). Therefore, the same kernel will apply to every location of the input map and the connections between two layers will share the same weights (thus weight sharing). Weight sharing is efficient because an object may appear in any part of an image and its appearance is invariant to its placement. Weight sharing reduces the number of free parameters in the network and makes it easier to train.  %It is sometimes called parameter sharing.

\textbf{Local and sparse connectivity}. Convolutional layers in CNN achieve  sparse connectivity by using a kernel much smaller than the size of input map ($h, w < 10$, usually).  Each node in a convolutional layer is only connected to a small number of nodes in the previous layer, \ie, a local region. This kernel is small because adjacent pixels and subregions of an image are more highly correlated than distant regions. 
%CNNs typically have one or a few number of fully connected layer whose nodes are connected to all nodes from the previous layer. 

%The key ideas of CNNs are compositionality and translation invariance. Namely, an object comprises a number of subparts which composites a holistic object. Lower layers of a CNN can capture the presences of subparts, which will be combined together at a higher layer. In addition, both objects and their subparts can be found in any part of an image. A CNN will handle part deformation and instance translation by applying the same set of filters to different locations, a scheme known as ``weight sharing.'' This also allows to reduce the number of tunable parameters significantly. 

%CNNs have a few unique components as follows.
\subsubsection{Nonlinear Layer}
Each convolutional layer is typically followed by a nonlinear activation function that applies to each element in the feature map. One of the most common activation functions is the rectified linear unit (ReLU):
\begin{equation}
f(x) = \max (0, x).
\label{eq:relu}
\end{equation}
This function will simply replace negative feature map values with $0$ and keep positive values. Other functions such as sigmoid or hyperbolic tangent function can be also used. The main advantage of the ReLU is that it runs much faster than those functions. 

Nonlinearity of visual models is important as it allows to capture a complex data distribution. Especially, nonlinear layers are essential in  deep networks because consecutive layers of linear operations will be simply collapsed into one linear layer. Thus, there will be no benefit of adding more layers to the network without nonlinear functions.  

\subsubsection{Pooling layer}

Pooling is another important operation in convolutional neural networks since it reduces computational complexity. A pooling layer takes an input feature map from the previous layer and generates a transformed map whose size differs from its input size. Most images and feature maps in a CNN are spatially correlated: values in closer pixels or nodes tend to be more similar because most scenes are continuous.  
Instead of keeping similar values redundantly from adjacent locations, one can simply choose the maximum response (or the average value) in each spatial neighborhood (pooling window) to represent the area. 
%This redundancy may be unnecessary if the goal is simply to determine the presence of a certain feature or an object in an image. In this case, the maximum response in each spatial neighborhood (pooling window) can be a concise sufficient statistic representing the area. 

\begin{figure}
\centering
\includegraphics[width=0.4\textwidth]{{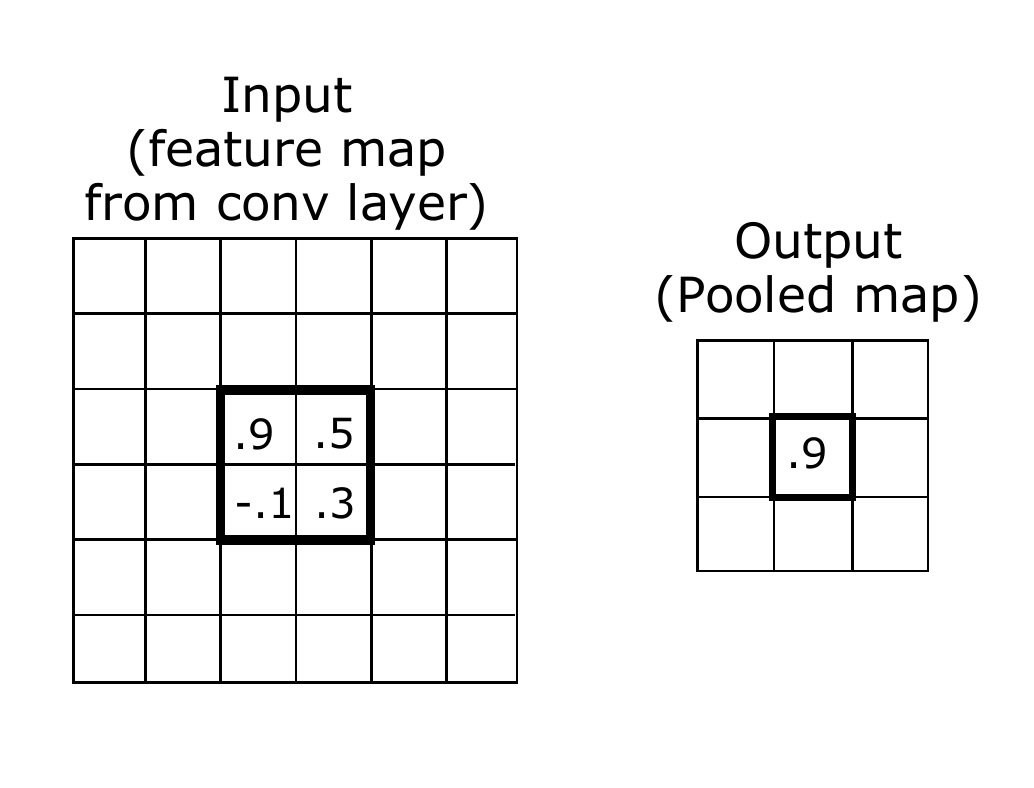}}
\caption{Illustration of a max-pooling operation of the window size $2 \times 2$. For each window, only the maximum value will be retained. 
}
\label{fig:maxpool}
\end{figure}

Specifically, a max pooling layer compares values in each sub window (e.g., a $2 \times 2$ window of pixels) of the input feature map and chooses the maximum value (see Figure~\ref{fig:maxpool}). Only these maximum values will be stored in the output map; the other values are disregarded. Removing non-maximum values also means that the resulting feature map will be of a smaller size than the input map. For example, an input image of size 256 $\times$ 256 will be downsized to 16 $\times$ 16 after applying 4 max-pooling layers of size 2 $\times$ 2.

%Since each sub window will have only one value after pooling, it is unnecessary to keep the original spatial resolution. A resulting feature map thus typically has a smaller spatial resolution according to the size of the pooling window (25\%, in Figure \ref{fig:maxpool}). Since CNN architecture usually has multiple pooling layers, and some convolutional layers also downsize the input by downsampling, an input image of a higher resolution will be repeatedly resized to a much smaller map (e.g., $7 \times 7$).

Pooling not only reduces the number of free trainable parameters but also helps the network achieve translation invariance, which is an important property for computer vision systems. One main difficulty in visual learning is high geometric variations of objects and parts arisen from part movements and viewpoint changes. 
%Objects consists of many individual parts which can move independently.  This movement will result in a geometric variation under which several images of the same object may look different. 
Robust computer vision system needs to handle such a geometric variation, and pooling operations help by disregarding small spatial perturbations within the pooling window.

\subsubsection{Fully connected layer}
\label{sec:fclayer}
CNN architectures used for classification include one or a few fully connected layers at the last stage. By definition, a fully connected layer densely connects all the nodes from the previous layer to all the nodes in the current layer. A convolutional layer encodes local information tied to specific image subregions distributed over a two dimensional map, through a sparse connectivity (\ie, nodes are selectively connected in a convolutional layer). A fully connected layer collects local evidences from all the subregions, captured in the prior convolutional layer, and outputs the overall likelihood of a visual concept which the whole network attempts to model. 

In the case of classification, the fully connected layers in a CNN are usually followed by a softmax function, which normalizes the final classification scores over categories. See Section~\ref{sec:imageclass} for details. 

%and merges together this scattered information before final classification.

%A fully connected layer densely connects all the units from the previous layer to all the units in the current layer.  This is because, for example, we want to classify whether an image contains a car or not but do not care where in the image it is. The fully connected layer will therefore collect evidence from all different subregions, captured in the convolutional layer, and output the overall likelihood of the presence of a car in the image. 

\subsection{Why Does It Work Well?}
Artificial neural networks have a long history in machine learning and computer vision and became extremely popular after  \cite{krizhevsky2012imagenet} demonstrated an impressive image classification performance on a benchmark dataset, Imagenet \citep{deng2009imagenet}. 

\begin{figure}[h]
\centering
\includegraphics[width=0.85\textwidth]{{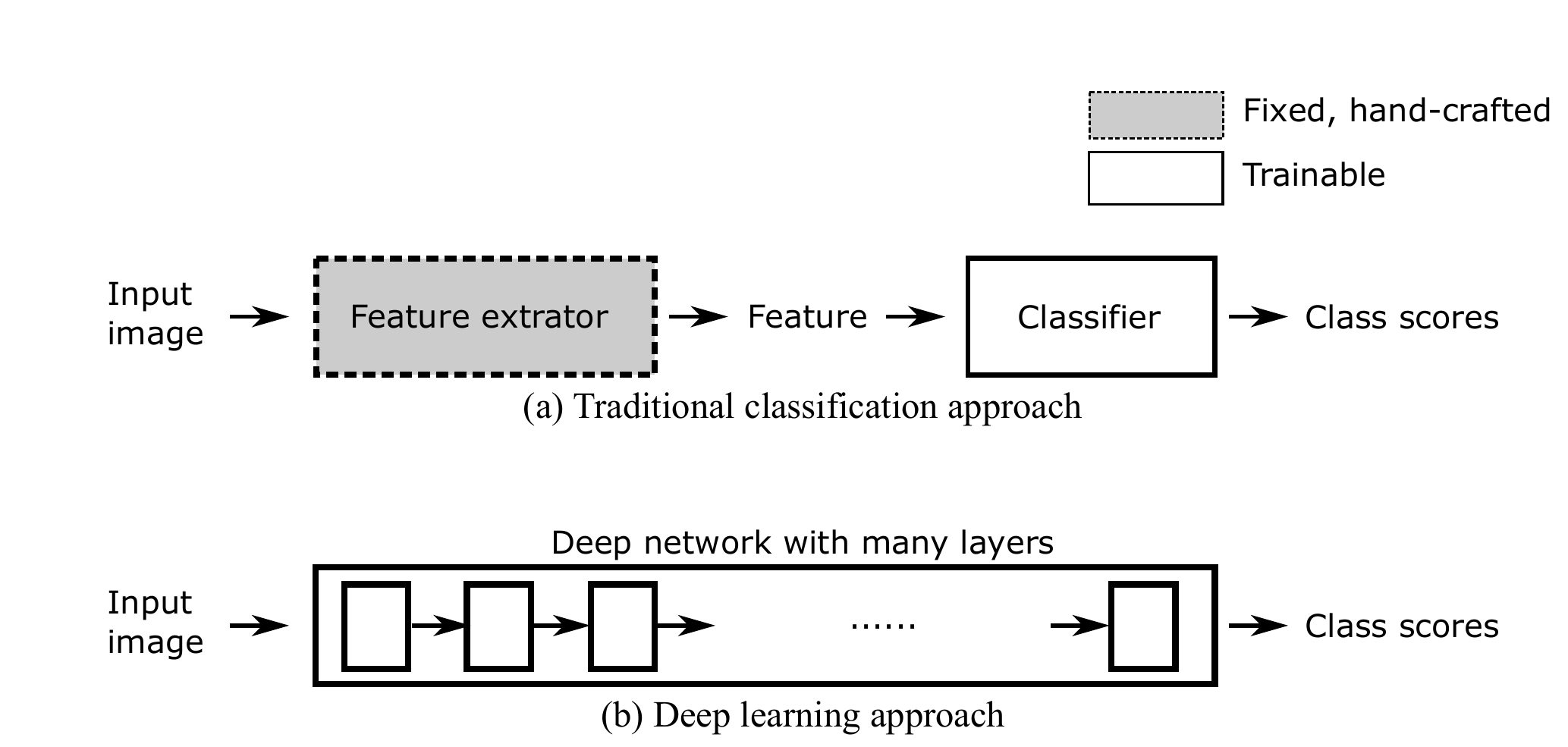}}
\caption{Comparing Deep Learning to Previous Computer Vision Methods}
\label{fig:nnidea}
\end{figure}

Traditional machine learning methods utilize a two-step process as shown in Figure~\ref{fig:nnidea}. Given raw input data (\eg, images), these methods first extract features (\eg, an edge histogram) using a fixed, handcrafted feature extractor.\footnote{The notion of handcrafting means that a researcher manually designs and defines the feature extraction function based on his insight. } These features should capture the most important cues in the raw data, which will be exploited by an off-the-shelf classifier in the second step. 

In contrast, deep learning based methods learn their representations directly from data without separate, hand-crafted feature extraction. 
%The most unique feature of deep learning based methods among other machine learning approaches, in contrast, is that their representations are directly learned from data. 
These methods employ a data driven approach in feature learning and train an integrated model that will automatically learn and capture both low- and high-level representations of data \citep{lecun1998gradient}. This approach is advantageous because the learning algorithm can discover many subtle features which are specific to the given task. In other words, the features in deep learning are optimized for the task during training, whereas in traditional methods, handcrafted features are invariable bottlenecks. It is also very common that \textit{all} trainable parameters in a neural network are \textit{jointly} optimized simultaneously, called end-to-end training. 

In addition, deep models contain a large number of layers, using the representation in lower layers as building blocks for the representation in higher layers. This idea of compositional and hierarchical modeling of  visual data has long been a key principle in computer vision and pattern recognition \citep{bienenstock1997compositionality,  felzenszwalb2010object}. Nevertheless, deep learning dramatically extends the number of layers possible by an effective learning algorithm \citep{lecun1989backpropagation}. 
%zhu2007stochastic

%Section~\ref{sec:training}
%Furthermore, many studies have shown that deeply learned features generalize better \textbf{(CITE)}, not only between different data samples within the same task but also between very different datasets across multiple tasks. \textbf{(REFER TO FINE TUNING)} These all suggest that it is imperative to learn features directly from data to achieve better performance. 

\section{Tasks in Computer Vision}
\label{sec:tasks}
This section describes common tasks that computer vision and deep learning techniques can solve. Following this section, Section \ref{sec:trainValidate} explains how to build and evaluate a CNN for these tasks and Section \ref{sec:appPolisci} discusses how these methods and tasks can be applied to political analysis. 
%, and Section \ref{sec:appEventsData} demonstrates with an application to event data.

\subsection{Image Classification}
\label{sec:imageclass}
One of the most well known research problems in computer vision is image classification. Given an input image, $I$, the goal of image classification is to assign a label, $y$, in a predefined label set, $Y$, based on the image content:
\begin{equation}
y^{*} = \arg\max_{y \in Y}{ p(y | I) }.
\end{equation}

In case of binary classification, $Y$ = \{positive, negative\}. In general, $Y$ may contain any arbitrary number of possible labels, \eg, generic object categories. The posterior probability for each category is computed upon a given input image and the classifier chooses one category with the highest output score. 

\begin{figure}
\centering
\includegraphics[width=0.9\textwidth]{{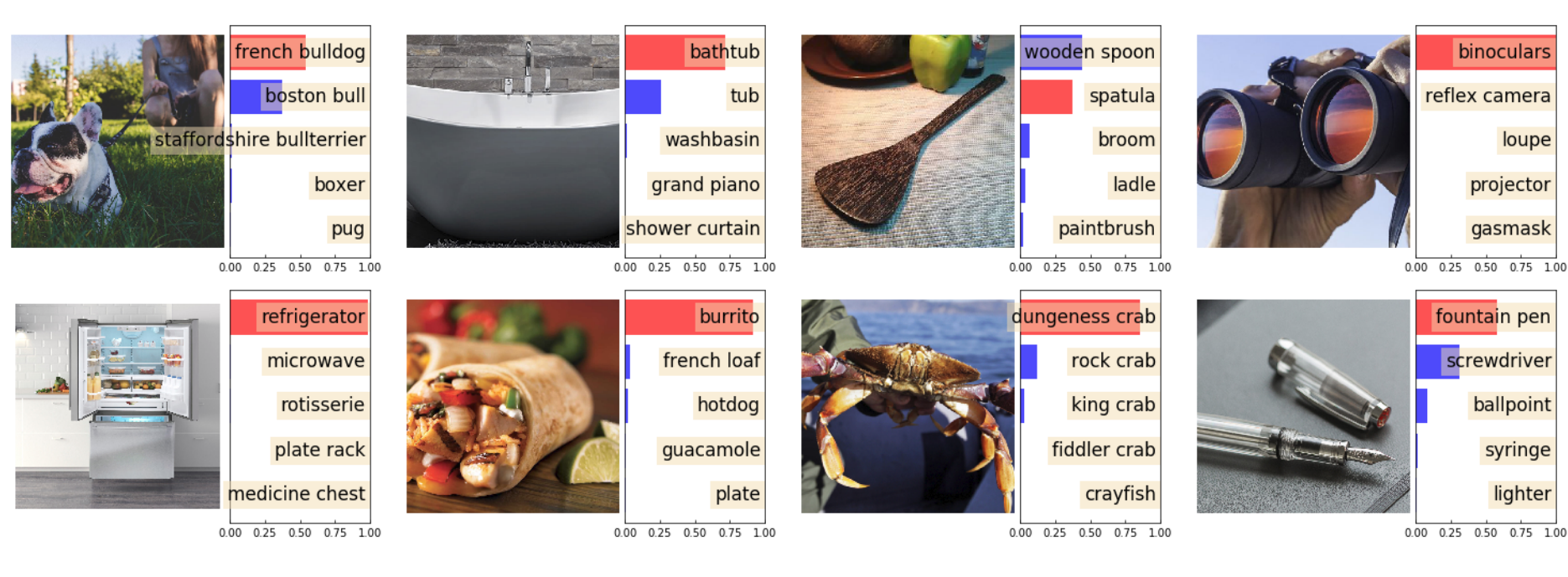}}
\caption{Example results of image classification with the confidence scores computed from a CNN.  Red color indicates the correct category and blue color indicates the incorrect categories. 
}
\label{fig:imagenet}
\end{figure}

One famous example of image classification is the ImageNet Challenge \citep{deng2009imagenet}, which is a public competition among diverse classification systems which are trained on and tested by identical image data and annotations. In this data, each image has been manually labeled with one category out of 1,000 categories. The categorization is object-centric and the correct label corresponds to the main object in each image. These categories include animals (\eg, bear, horse), vehicles (\eg, school bus, scooters), musical instruments (\eg, piano, acoustic guitar), and so on. 

The CNN architecture depicted in Figure~\ref{fig:cnn} can be used for image classification. The softmax function is commonly used in multiclass classification to normalize output scores over multiple categories such that the final scores sum to 1 as a probability distribution. Suppose that the last fully connected layer outputs a vector $\mathbf{x} = (x_1, x_2, ..., x_n)$, where $x_k$ is the raw output score before normalization for the $k$-th class out of $n$ classes. The final score will be obtained as follows. 
\begin{equation}
p(y = k | I) = f_k(\mathbf{x}) = \frac{\exp(x_k)}{\sum_{j=1}^n\exp(x_j)}.
\end{equation}

In the case of multilabel classification, each class is classified separately and an image is allowed to be assigned more than one positive class. For instance, an image may contain a horse and a dog. This image will be labeled with both classes in multilabel classification. In this case, the softmax function is replaced with sigmoid functions applying to each output node separately because there will be no normalization over classes. 

\subsection{Object Detection}
\begin{figure}
\centering
\includegraphics[width=0.95\textwidth]{{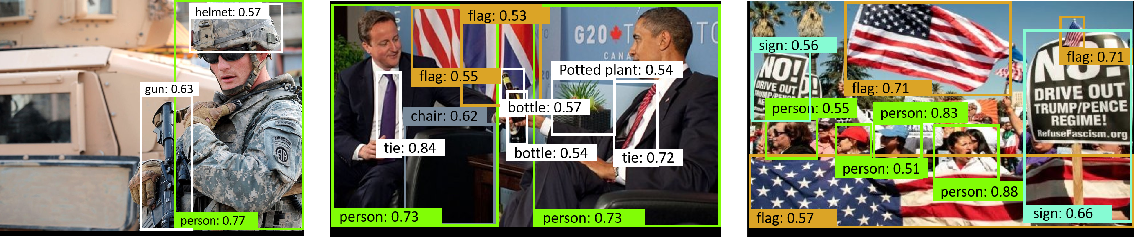}}
\caption{Example results of object detection. }
\label{fig:objectdet}
\end{figure}

The goal of object detection is to localize object instances and classify their categories in a given input image. The output of object detection is a set of detected object instances including their locations and categories. Figure~\ref{fig:objectdet} shows example results of object detection with detection scores. 

%each of which is represented by a tuple of object location and category. Object location is commonly represented by a bounding box, which is a rectangle of minimum size that covers the detected instance. 
%where x, y, w, h indicate the coordinates and size of the bounding box. The object class is similar to the image class defined in the previous subsection and corresponds to one of predefined object categories in advance. 

Object detection is considered a more complex problem than image classification because the model should classify the types of object instances and their locations. In practice, many object detection systems utilize a two-stage procedure. First, the system generates a number of generic object ``proposals'' from an input image \citep{uijlings2013selective}. 
%zitnick2014edge
These proposals are image subregions which the system believes are likely to capture an object instance, regardless of its category. Second, the image classification step will then apply to each object proposal to determine its category or reject it as a background region. In more recent methods \citep{ren2017faster}, these steps are integrated within one model, yielding a better accuracy and computational efficiency. 

An object location is represented by a rectangular bounding box, $(x, y, w, h)$, indicating the coordinates and the size of the bounding box. This bounding box is the rectangular area of the minimum size that can cover all the pixels that the object occupies in the image. 

%While bounding boxes are a compact representation for object location and size, they also contain background pixels. In case it is necessary to separate only foreground pixels from background regions, one can use semantic segmentation to assign a class membership to each pixel in an image \citep{long2015fully, chen2018deeplab}. 

%Object detection can also apply to many real world applications. For example, self-driving cars can detect pedestrians or other cars and avoid collisions using object detection from camera inputs along with other sensor data. Another application of vehicle detection is automated license plate detection and recognition for toll collection. 

\subsection{Face and Person}
The human face has received enormous research attention as a special domain in computer vision since the 1970s for two main reasons. First, it has many useful applications, \eg, a security system or a survey tool. Second, it is relatively easier to handle face images compared to other objects, because the appearance of a human face is consistent across individuals but distinct from other objects. These properties motivated early innovative explorations in the topic such as automated feature extraction \citep{Kanade-1977-15584}, feature learning with neural networks \citep{fleming1990categorization}, and classification based on statistical analysis of data \citep{belhumeur1997eigenfaces}. Existing work in this topic can be categorized into three areas: face detection, face recognition, and face attribute classification. 

\textbf{Face Detection. }
Face detection refers to finding the location of every face in an input image. It can be posed as a binary classification problem where the classifier is required to determine whether each image sub-region contains a human face or not. There are mainly two approaches in the general objection detection task: the sliding window and the object proposal-based search. The sliding window is an exhaustive search algorithm which simply examines every possible ``window,'' i.e., 2-D rectangular sub-region of an image. This method will extract the feature from each window and classify its label based on the feature description. This can achieve the best recall but is inefficient because it has to examine all possible windows. In contrast, the proposal-based search is a selective procedure which first selects a subset of windows by rejecting a large number of easy negatives, \eg, black background, and performs subsequent classification only for the selected windows, \ie, object proposals. 

\cite{viola2004robust} proposed the most famous face detection algorithm, commonly known as the Viola-Jones detector.  This system employs ``haar-like'' features, which measure local contrast in image intensity, and evaluates the weighted sum of a number of local feature values. Adaptive Boosting is used for selecting the best features from a larger feature pool and determining the optimal weights.  It combines a number of ``weak'' classifiers to construct a strong one by iteratively adding the best weak classifier at each round and adaptively adjusting the weight of each sample in training data.
%such that the examples misclassified by the previous classifier should contribute more at the present round. They also utilize a cascade in classification; the model consists of several stages and input windows are tested through these stages sequentially while a number of negative windows are rejected at each stage. \\

\textbf{Face Recognition and Verification. }
Face recognition is a task to classify the identity of a person from a facial image and face verification is a task to compare two facial images and judge whether they are the same person. They are usually based on the same face model which takes an input facial image and computes the facial feature from it; face recognition can build on face verification by comparing one input facial image against every face in the database of people with known identities. 
%In order to improve the recognition accuracy, input facial images, which may capture faces in different orientations or viewing angles, are often transformed through facial alignment \citep{xiong2013supervised} or frontalization \citep{hassner2015effective}. 

Face recognition models are either part-based or holistic. In part based approaches, different facial regions, such as forehead or mouth, are detected and modeled separately, and the local features from multiple regions are combined for final classification \citep{kumar2011describable}. In holistic approaches, the appearance of a whole facial region is directly modeled \citep{belhumeur1997eigenfaces}. 
%turk1991face

Most recent approaches in face recognition are based on convolutional neural networks in which local parts are not explicitly defined but implicitly captured in the model hierarchy. A recent study by Facebook \citep{taigman2014deepface} has reported that their model based on a CNN is as accurate as human annotators in face verification after trained from 4.4 million labeled face images obtained from their users. 
%While these images are not publicly available, there exist public large scale face image datasets that can be used for training robust models \citep{guo2016ms}. \\

%Traditional approaches in face recognition or verification used to model different facial regions such as forehead or mouth separately and combine the local features together to perform the final classification \citep{kumar2011describable}. 

\begin{figure}
\centering
\includegraphics[width=0.95\textwidth]{{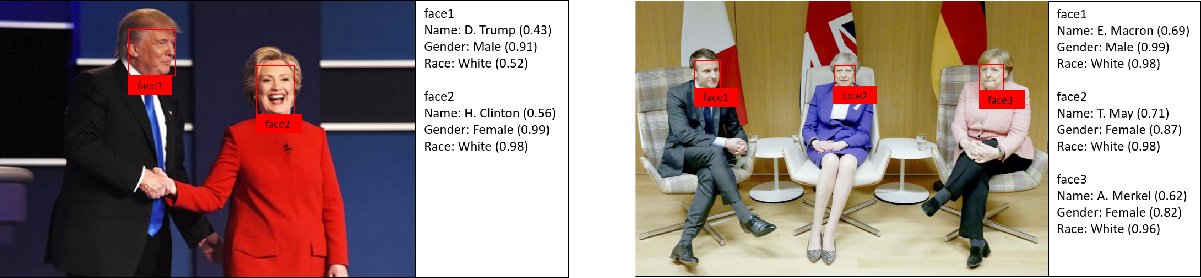}}
\caption{Example results of face detection, recognition, and attribute classification.}
\label{fig:face}
\end{figure}

\textbf{Human Attribute Classification. }
A face provides clues for recognizing demographic variables (\eg, gender, race, age), emotional states, expressions, and actions, commonly referred to as human attributes in computer vision. Figure~\ref{fig:face} shows two example results of face recognition and gender and race classification from facial appearance. Large scale datasets of facial images and attribute annotations are also available \citep{liu2015deep} and enable training a deep CNN with a similar structure to an image classification model. 
%This model should have the same number of output nodes in the output layer as the number of attributes that one wishes to train. Each attribute will be modeled and classified separately.

%While faces are usually the most reliable source for human attribute classification, other body cues, such as clothing, can further enhance the accuracy of human attribute classification. Some methods therefore attempt to detect body parts, extract evidences from local regions, and merge multiple cues together to make a final decision \citep{joo2013human}. Such an approach is also helpful when a face is occluded or not visible.

\section{Training and Validation}
\label{sec:trainValidate}
This section explains how one develops an image classification system using CNN, focusing on model training and validation. We will discuss practical issues in training a model and introduce tools to diagnose the model performance.  
%To help readers understand the method better, we will also present concrete examples and analysis from a recent study of protest event analysis \citep{won2017protest}. 

\subsection{Training}
\label{sec:training}
As in other machine learning methods, training means estimating from training data optimal values for model parameters that minimize a target loss function, \eg, an error function or a learning objective of the model. In CNNs, these parameters include kernel weights in convolutional layers (Section~\ref{sec:convlayer}) and weights in fully connected layers (Section~\ref{sec:fclayer}). 

There exist many loss functions widely used in training a CNN. One can select a specific loss function or a combination of multiple loss functions depending on the task and the output dimension. In image classification, for example, the most popular loss function is cross-entropy loss, also called log loss. In a binary classification task, the binary cross-entropy loss is: 
\begin{equation}
loss_{\text{bce}} (y, \hat{y}) = - (y \cdot \log{\hat{y}} + (1 - y) \cdot \log{(1- \hat{y})})
\end{equation}
where $y \in \{ 0, 1\}$ is the true label for the example and $\hat{y} \in (0, 1)$ is the output value computed from the model. In training, all the model parameters (model weights) are optimized to minimize this loss function across the entire training set. 

This optimization is performed by stochastic gradient descent with the backpropagation algorithm \citep{lecun1989backpropagation}, alternating between forward and backward passes. In the forward pass, given an input (an image), the network evaluates the model outputs (classification results) and computes the loss function based on the ground truth output labels. In the backward pass, the gradient of the loss function is propagated backward by the chain rule and the model weights are updated accordingly. 
%It is called backpropagation because the gradients are computed from the last layer of the network to the first layer.

Training a CNN with many layers can take several weeks to several months, even with a GPU. Stochastic gradient descent is an iterative procedure and updates the model parameters incrementally through many iterations. Training deeper models requires a larger training set and more iterations than shallower models. 

\textbf{Fine Tuning.} Fine tuning is a popular technique which can accelerate the training process of neural networks. When training a network, all the weight values are typically initialized to small random values at the beginning and then gradually updated. Instead of using random values, one can also take the weight values from an existing model which was already fully trained from another dataset and initiate a new training process. This procedure is called fine tuning as an existing model is to be tuned to another task. This existing model is called a pre-trained model. For example, one may use a pre-trained model trained for face detection to initialize the weight values of a new model for person detection. 

The underlying idea is that CNNs, especially in their lower layers, capture features that can transfer and generalize to other related tasks as these features can be shared across the tasks, \ie, transfer learning. In visual learning, these sharable representations include elementary features such as edges, color, or some simple textures (Figure~\ref{fig:deconv}). Since these features can commonly apply to many visual tasks, one can simply reuse what has been already trained from a large amount of training data and refine the model to the new data. 

Fine tuning is also helpful when the training data is insufficient to train a deep network. Training a deep model requires a huge amount of training data when starting from scratch, but such a large dataset may not be available. In the case of fine tuning, the pre-trained model was already trained on a large dataset and can provide a robust starting point for the new task. 

%at least tens of thousands of labeled images, a limitation in many situations. Instead of this many, or more, examples, one can start from a pre-trained model and fine tune it with a much smaller dataset.  The result is more robust than training from scratch on a smaller dataset of images.  
%Fine tuning is sometimes referred to as transfer learning.

\subsection{Validation and Interpretation}
Deep neural networks, despite their remarkable performance, often receive criticism due to the lack of interpretability of their results and internal mechanisms compared to simple models with a handful of explanatory variables (linear classifiers, for example). A deep model typically comprises millions of weight parameters (edges between nodes), and it is impossible to identify their meanings or roles from the output. 

Once a model is trained, one can measure its generalization error using a validation dataset which does not overlap with the training set. As in other classification problems, the performance, \eg, accuracy or goodness, of a CNN-based classifier can be measured by several metrics, including raw accuracy, precision and recall, average precision, and many others. These measures, however, do not explain how the model achieves such an accuracy. 

To make the result of deep models more explainable and interpretable, several methods have been proposed. 

\subsubsection{Language-based Interpretation}
As humans use language to explain a concept, one can develop a joint model that incorporates visual and textual data such that the text part explains its visual counterpart. For example, image captioning generates a sentence describing visual content in an input image \citep{kiros2014multimodal} or text-based justifications to explain why the model produces such outputs \cite{hendricks2016generating}. 

Another line of research on text-based interpretation of visual learning utilizes questioning and answering \citep{antol2015vqa}. Such methods take both an image and a text question as input and output a text-based answer to the input question. This allows a more flexible interface between a user and a model than a traditional classification task, which essentially asks a fixed question to the model. 

The key limitation of these methods is that they do not generalize: they are unable to deal with novel content or questions. The models are trained on image-text pairs and simply reproduce the mapping learned from the training data. When the model is given a novel question which was not given during training, it may not understand the meaning of the question. 

\subsubsection{Visualization}
Another way of understanding how a deep network produces its output is through visualizations. Since convolutional neural networks are largely used for visual learning from images, it is especially effective to visually illustrate their mechanism so that the user can understand it better. We introduce the two most popular approaches: feature-based and region-based. 

\begin{figure}
\centering
\includegraphics[width=0.8\textwidth]{{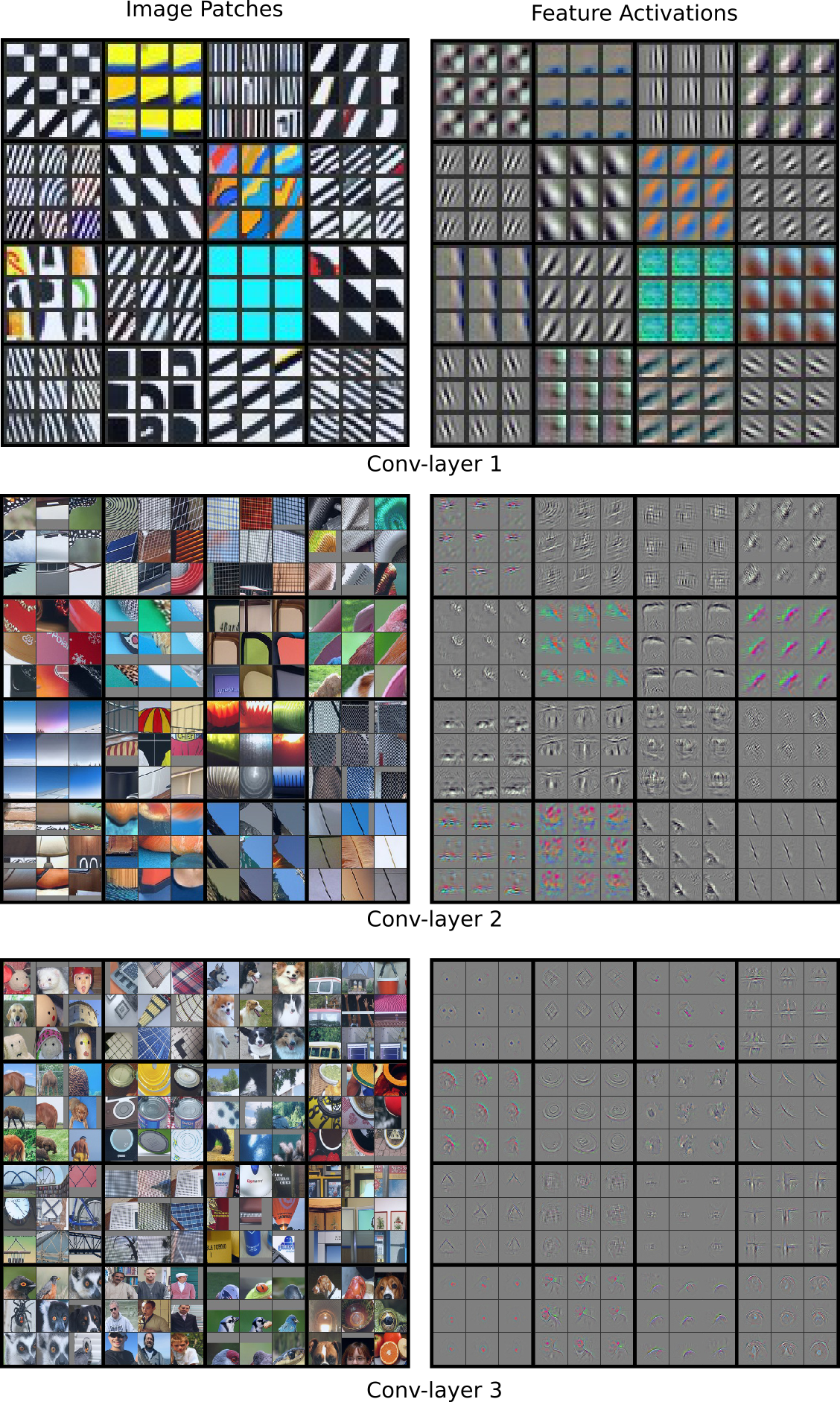}}
\caption{Visualization of feature activations at different layers in a CNN by a deconvolutional network \citep{zeiler2014visualizing}. For each layer, the left panel shows groups of similar image patches which produce high activation values for a specific node in the layer. The right panel shows corresponding feature visualizations.  }
\label{fig:deconv}
\end{figure}

Figure~\ref{fig:deconv} provides examples of the feature-based approach.  This approach uses a ``deconvolutional'' network \citep{zeiler2014visualizing}, which is akin to a reverse CNN. Unlike a CNN which collects feature activations at multiple layers to make a final output, a deconvolutional network redistributes the contributions of each feature and projects their importance back to the input pixel space.  Figure~\ref{fig:deconv} shows that visually similar image patches that contain the same image feature (left sub-panel) will trigger high activation scores in the same node in the network that captures the image feature. The image feature can be visually identified from the feature activation maps (right sub-panel). Moreover, this visualization also confirms that the lower layers in a network respond to the low level visual similarity such as color or texture, and the higher layers tend to capture semantic similarity at the object category level.

\begin{figure}
\centering
\includegraphics[width=0.8\textwidth]{{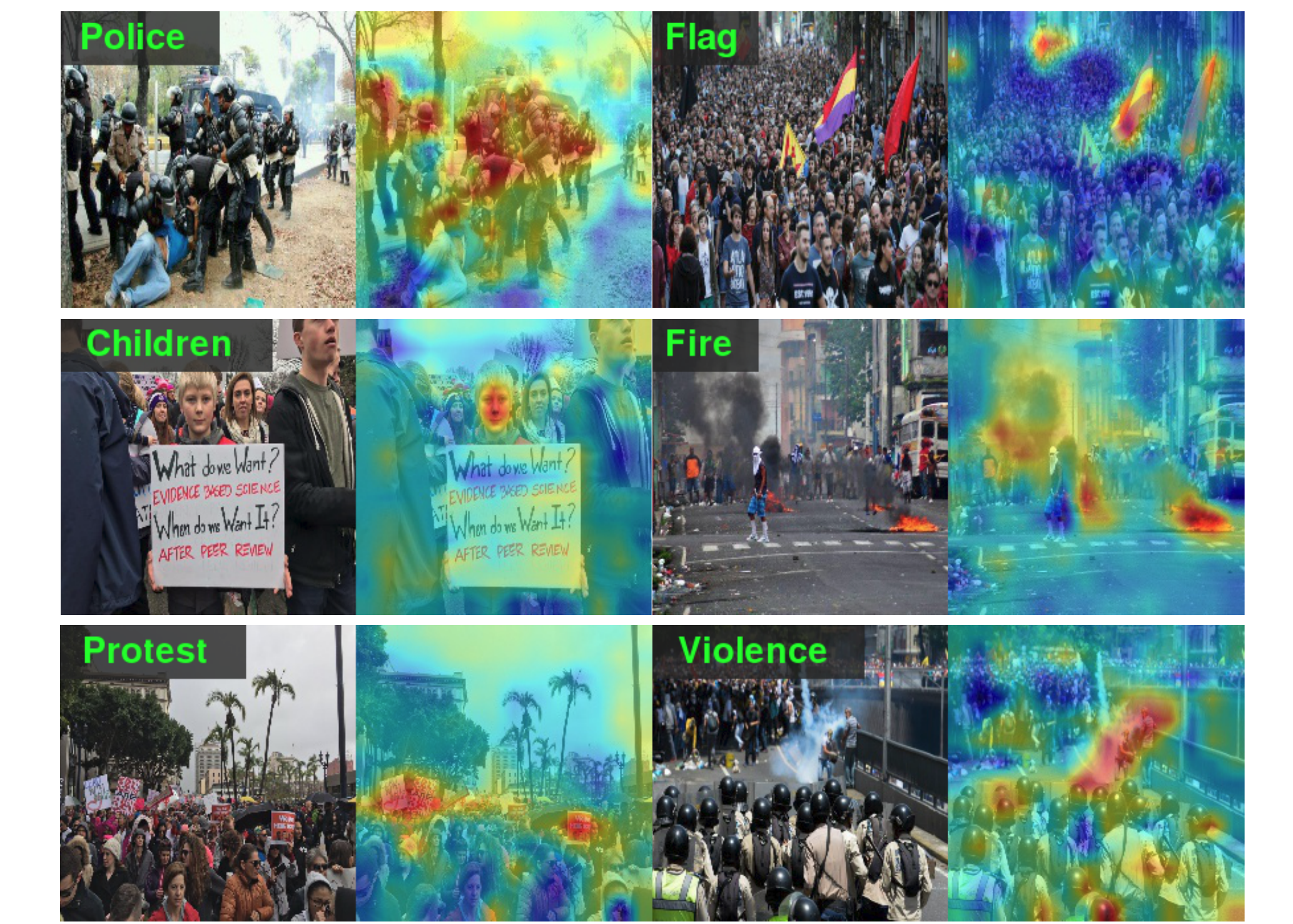}}
\caption{Visualization of region importance to visual concept classification by Grad-CAM \citep{selvaraju2017grad}. Examples are taken from a recent protest image analysis \citep{won2017protest}. Important regions are marked by red color. }
\label{fig:gradcam}
\end{figure}

%zhou2016learning
 Figure~\ref{fig:gradcam} shows the region-based approach, highlighting important regions which more contribute to the model output \citep{selvaraju2017grad}. This example visualizes the region importance of a CNN trained for protest image classification \citep{won2017protest}. This network classifies whether an image contains protesters, police, or fire, and it estimates the level of perceived violence, using the Grad-CAM method \citep{selvaraju2017grad}, which evaluates a weighted sum of feature response maps from all convolutional layers in a network, to highlight important regions by color. 
 %While these methods do not provide fine-grained feature details, they do indicate important features. 
 For example, Figure~\ref{fig:gradcam} shows that abstract concepts, such as protest and violence, are classified from less abstract concepts such as individuals holding signs or the presence of smoke. Visualization helps users understand how complex classification is internally performed.

\section{Applications in Political Science Research}
\label{sec:appPolisci}
This section describes existing research in political science relying on visual data. Since automated visual analysis is still very new, few works in political science have adopted the methods. We will therefore also discuss existing manual analysis on visual data and explain the potential utility of automated methods to different domains. 

\subsection{Political Behavior}
\label{sec:DemoStudies}
Visual data can identify demographic information about a person such as gender, race, age group, or other features based on facial appearance. A few recent studies have used profile images of individual users in social media in order to infer such demographic information about the users. \citet{wang2016deciphering} analyzed demographic compositions of the followers of Donald Trump and Hillary Clinton in Twitter in the U.S. 2016 presidential election, using profile images. A similar approach was also used to analyze promoter demographics in various social and political campaigns in Twitter \citep{chakraborty2017makes}. Both studies used the same commercial software to classify demographic attributes of people from photographs \citep{faceplusplus}.  

Using deep learning on images holds much promise in extending our knowledge of the relationship between demographics of protesters and policy change, an area that has received little direct testing \citep{Fisher2005}.  For example, a 1986 study of 1964 Freedom Summer participants uses survey data of college participants collected twenty years prior  \citep{McAdam1986}.  A study of participants in the 1989 East Germany revolution asks respondents their age, gender, marital status, number of kids, and education \citep{Opp1993zzz}.  Others conducted a survey of protesters in Egypt's Tahrir Square, to measure how social media use varied by gender, age, and education \citep{Tufekci2012}.  A study of participants in Ukraine's 2004 Orange Revolution finds participants came from diverse ideological backgrounds, with different demographic categories have opposing effects on the probability of protesting \citep{Beissinger2013}.

These studies each focus on one protest event, making it difficult to generalize about how gender, age, and race affect protest size.  For example, \cite{Beissinger2013} and \cite{Opp1993zzz}  find older individuals are less likely to protest, but \cite{Tufekci2012} find the opposite.  \cite{Opp1993zzz} find men less likely to participate than women, the opposite again of \cite{Tufekci2012} and \cite{Beissinger2013}.  While these conflicting findings could be due to various factors - survey design, political regime effects, and differing economic opportunities, to name a few - another possibility is that scholars have not been able to pool demographic correlates in models.  Because of the difficulty of measuring demographics, research focuses on one or two protest waves, uses surveys, and data are usually gathered after the fact.  The ability to generate demographic data about protesters using deep learning and computer vision may permit a more stable identification of which individual features correlate with protest participation. 

%Better measuring protester demographics - measuring more protesters across more protests - will also help scholars understand the relationship between protester characteristics, political institutions, and policy change. For example, a politician should also be more responsive to protests when their demographic composition closely matches their district's \citep{Downs1957}.   On the other hand, a Senator may be less responsive to protesters than a Representative, since the former have 6 years, versus 2, between elections.  How institutional factors interact with individual characteristics 

\subsection{Political Communication}
Political communication studies the interaction and communication among politicians, media, and the public across speeches, public debates, newspaper articles, and television broadcasts. 
%Prior to digital technologies, the quantity of political messages was relatively small and generated by a small number of politicians and news outlets.  A substantial portion of such data, however, was not archived or accessible to researchers. In contrast, 
Recent advances in technology, especially the rise of social media, have prompted the creation and transmission of an enormous amount of political communication data.  A substantial portion of such data is accessible to researchers, and scholars have developed automated machine learning-based techniques for analysis of political text data \citep{Grimmer2013}. These techniques allow researchers to discover latent topic structures from an unstructured document set \citep{grimmer2010bayesian, roberts2014structural} or measure opinions or sentiments of authors from text \citep{tumasjan2010predicting, o2010tweets}.

Political scientists have also paid close attention to the visual dimension of political communication 
\citep{gilliam2000prime, barrett2005picture, rosenberg1986image, grabe2009image, barnhurst1997image, hansen2015images, schill2012visual}. 
%\citep{gilliam2000prime, powell2015clearer, barrett2005picture, rosenberg1986image, grabe2009image, barnhurst1997image, hansen2015images, barnhurst2012political, schill2012visual}. 
Most people access news through multimodal media; even newspapers devote significant space to photographs. Presidential debates, for instance, may be seen as an event mainly about verbal exchanges between candidates. They are, however, televised to the viewers, with many visual cues which communicate emotions and tensions between them to the viewers \citep{shah2016dual}. Indeed, several studies argue the nonverbal cues and visual exposures of politicians in media may encode their emotions and invoke voter reactions \citep{sullivan1988happy, grabe2009image, mchugo1985emotional}. 
%bucy2000emotional

Automatically accessing, collecting, processing, and analyzing visual data has been the main bottleneck preventing existing manual studies from scaling. To overcome such a difficulty, recent studies use computer vision models to analyze characteristics pertaining to the political dimensions of actors or events portrayed in large collections of images. \cite{Joo2014} demonstrated that an automated method can be used to infer the hidden communicative intents from photographs of politicians, highlighting certain personal traits to visually persuade the audience. The study detects several scene components such as facial expressions and surrounding objects and measures the visual favorability of politicians. The study shows that the visual favorability automatically estimated from newspaper photographs positively correlates with the public opinion of the politician. 
%Subsequent work has also shown that it is possible to improve the prediction accuracy by using deep learning models on the same dataset \citep{Huang2016}.

%Research in this area has commonly identified facial expressions of politicians as an indicator of overall favorability. For instance, a smiling face is more likely to convey a positive sentiment about the main person being depicted. Based on this assumption, \citet{groeling2016} have examined the degree of media bias present in TV news programs in the U.S. by analyzing facial expressions of presidential candidates across networks. This study utilized a CNN based facial attribute recognition model to classify diverse facial expressions.  Going beyond traditional professional sources, attempts have been also made to analyze political images in social media. For instance, \citet{you2015multifaceted} have analyzed multimodal cues of Flickr posts related to presidential candidates in the U.S., showing it is possible to predict election outcomes based on facial expressions and hashtags. 

Computer vision methods have also shown the potential effects of politicians' facial appearance on voters' trait judgment and election outcomes. Personality inference from facial appearance is a well studied topic in psychology \citep{zebrowitz2008social}, and political scientists have attempted to explain public responses and election outcomes based on physical appearance of political leaders such as their visually-inferred competence \citep{todorov2005inferences}. 
%ballew2007predicting
Automated models have been used to extract visual features from facial images to predict subjective trait judgments on various dimensions such as intelligence or trustworthiness \citep{rojas2011automatic, vernon2014modeling}. It has been also shown that automatically inferred facial traits predict the actual election outcomes \citep{Joo2016}. 

Computational approaches offer several important advantages over manual investigations. First, a manual study requires a large pool of participants for reliable coding, which makes experiments expensive and time consuming. Second, since the study depends on participants' subjective evaluation, it often yields inconsistent, conflicting results. Third, manual studies cannot exclude participants' prior exposure and knowledge about politicians. In contrast, computational approaches are inexpensive to execute, entirely reproducible, and transferable from ordinary, unknown people to prominent politicians. 

\subsection{Development}
Computer vision techniques also hold promise in broadening and refining measures of development by using remote sensing data.  ``Remote sensing'' refers to the passive gathering of auditory or visual data about a place using tools the researcher does not control directly.  Common examples are detecting animals using movement sensors or measuring forest destruction using spectral (image) data.  %While the tools and methods have been developed outside of political science, the use of 
Remote sensing data is of use to many research questions in political science, especially those that involve socioeconomic indicators.

Spectral data can measure different features of cities, such as the distribution of building types, as well as land use in rural areas \citep{Jensen1999}. Imagery with a resolution of one meter or smaller can provide data on socioeconomic characteristics as they vary by neighborhood, allowing for frequent census-like data creation, an ability especially useful in countries with no, or irregular, census \citep{Tapiador2011}.  For agricultural areas, it can measure changes in rainfall and crop growth, proximate measures of income for many countries \citep{Tote2015}.  Since income shocks are a precursor to civil conflict, data that accurately measure subnational changes in income could act as an early warning system \citep{Hsiang2013}.

A canonical example of remote sensing is using light emissions to measure wealth, which works even for small geographic units in poor countries \citep{Weidmann2017}.  Convolutional neural networks have recently been applied to publicly available satellite imagery to measure household consumption and wealth, verified using household survey data \citep{Jean2016}.  The same tools and data can further help measure various socioeconomic characteristics at the household, neighborhood, or city level.

%Any research question that requires, or would benefit from, socioeconomic characteristics at the household, neighborhood, or city level would therefore benefit from training a deep learning model on satellite imagery data.  

Image data also provide access to temporal changes in local regions. For example, a model that accurately recovers built features of towns and cities could provide insight into how institutions affect recovery from natural disasters.  If images exist of the same area immediately before and after a natural disaster, the physical and geographic extent of damage as well as the speed and amount of recovery may be measurable.  These dependent variables may then be related to various institutional independent ones.  Recovery may occur more quickly in democracies than non-democracies or in countries with free media, for example.  In democracies, subnational variation could depend on whether a disaster strikes a powerful politician's district or if there is an impending election.  %For example, Hurricanes Irma and Maria caused widespread destruction through the Caribbean and United States, and differential rates of recovery may map to intra and international variation in political institutions.  
%Given the advances in computer vision and machine learning, the chief difficulty is finding publicly available, longitudinal imagery data.

More broadly, it should be possible to measure socioeconomic variables using photographs of places taken by people.  Manual analysis of Google Street View (GSV) imagery shows that photographs of cities taken at random times as Google's vehicles map them recovers public health data in the United States \citep{Odgers2012,Wilson2012a}.  A model trained on GSV images recovers income by block in New York City \citep{Glaeser2015}, and a deep learning model of cars in GSV images can measure income, race, and education at the precinct level \citep{Gebru2017}. % Another promising approach is to pay people to take photographs of specific phenomena, such as the price of goods at a supermarket or the prevalence of anti-incumbent signs at a protest \citep{Premise2017}.  

\subsection{Subnational Conflict}
%Remote sensing data is especially promising where the measurement of variables across political units, intra or international, would be cost prohibitive.  For example, 

A persistent debate in the civil war and protest literatures is the extent to which participation is driven by economic (``greed'') or political (``grievance'') motivations \citep{Collier2004,Kern2011}.  Those two concepts are notoriously difficult to operationalize, and researchers rely on measures such as the availability of natural resources (greed) or aggregate economic statistics such as gross domestic product (economic grievance).  Because studies rely on third parties' reports of these variables and generating those measures ranges from difficulty to almost impossible, variables are aggregated geographically to the state or country level and temporally to the yearly level.

Using computational approaches, greed and grievance can be measured with more geographic and temporal precision.  For example, greed is measurable using the precise outline of diamond mines, virgin forests, or oil deposits, which can be observed from satellite data or resource maps \citep{no-extraction-without-representation}.  Grievance should be reflected in city-level variation in economic activity measurable using light emissions \citep{Weidmann2017}.  While developing deep learning models on a large corpus of images from cities or countries with very different built and natural environments is not trivial, doing so is impossible with public data.  Moreover, doing so might be the only feasible way of acquiring these data in countries without detailed socioeconomic data.  This approach is likely to be even more useful for studying antecedents of civil wars, since they occur largely in poor countries \citep{Fearon2003}.  
Images can also be used to measure state capacity.  Humans-as-sensors can take photographs of specific objects, such as produce in a market, road conditions, or school conditions, using smart phones \citep{Premise2017}.  These images can give disaggregated information about a state's ability to repress intranational conflict, as well as the ability of rebels to attack the state.  Maps are also images, and digitizing them can provide historical data on state capacity, especially power projection, that current measures, such as GDP, may not capture \citep{Hunziker2018}.

Applying deep learning models to images would be especially useful to scholars researching state repression and protest.  For example, research into the repression-dissent puzzle consistently finds inconsistent results.  Repression may have no effect on dissent \citep{Ritter2016}, increase it \citep{Francisco2004,Steinert-Threlkeld2017}, decrease it \citep{Ferrara2003}, or have time varying effects \citep{Opp1990,Rasler1996}.  Data on the severity of violence, size of crowds, and initiator of violence that varies by city and day could provide a more definitive answer to these dynamics.  For an example of what these data would look like, see \cite{won2017protest} and \cite{STZ2018}.  For examples of work that automatically code protest data from images, see as well \citet{Torres2018} and \citet{Zhang2018}.\footnote{\cite{Cowart2016} manually evaluates images of Black Lives Matter protests newspapers and broadcast media tweeted.}  As of this writing, \citet{won2017protest} is the only one that is published. 
%Another example of visual analysis of protests in South Korea and Hong Kong is also provided in the Supporting Information. 

Finally, image data can generate data useful for interstate conflict research as well. For example, convolutional neural networks can detect fixed surface-to-air missile launchers eighty times faster than humans with the same accuracy \citep{Marcum2017}.

\section{Demonstration: Protest Analysis with Images}
As a demonstration of the method we discuss in the paper, we present an example analysis which can show the potential to use images as data to study details of protests that have eluded text-based datasets. It focuses on measuring crowd size, violence, and the demographic composition of protesters, which is fundamentally not possible using text to generate event data.

Specifically, we focus on measuring various features of two protests.  The first is the 2016-2017 protests in South Korea against President Park Geun-hye.  Revelations in October 2016 that President Guen-hye received council from a Rasputin like figure triggered large protests, and those protests persisted through her impeachment on March 10, 2017.  The second is the 2014 Hong Kong protests against changes to Hong Kong's electoral system seen as contradicting the ``One Country, Two Systems'' relationship with China.

We have developed a pipeline that identifies faces in a photo and estimates each face's sex (male or female) and race (white, black, or Asian)\footnote{We focus on the cases of South Korea and Hong Kong in this paper where the majority of the population are Asian and thus omit the variable race in analysis.}.  It also identifies if an image contains a child's face.  We can also measure whether a protest image contains police or fire; whether protesters are holding images; and the amount of violence in a protest image.  Section \ref{crowdsize} shows that computer vision and deep learning applied to geolocated protest images shared on Twitter accurately recovers protest size in South Korea and the United States.  (We do not analyze the United States in more detail because its protests have not lasted more than one day.  We do not analyze Hong Kong because we could find no other source of size information.)  Section \ref{skprotests} shows that this approach can measure daily level changes in the composition of protesters and violence at protests; these daily measures can then be used as variables in a regression. 

\subsection{Measuring Crowd Size}
\label{crowdsize}
Reliably measuring crowd size is an open problem, and estimates for protest crowd size are not consistently available.  Estimates that do exist come from state authorities or protest organizers, and systematic academic datasets of those estimates occurs when newspaper articles provide those estimates.  The only academic dataset that provides crowd size estimates for a protest in our study is the Crowd Counting Consortium \citep{Chenoweth2017}, and it only documents United States protests.  The other major dataset that provides crowd size estimates is ACLED, but it focuses on Africa, the Middle East, and South Asia and is updated with a delay.    

\begin{figure}[h]
\centering
\caption{Summing Number of Faces Accurately Measures Protest Size in the United States}
\label{img:facesVerify}
\includegraphics[width=.65\linewidth]{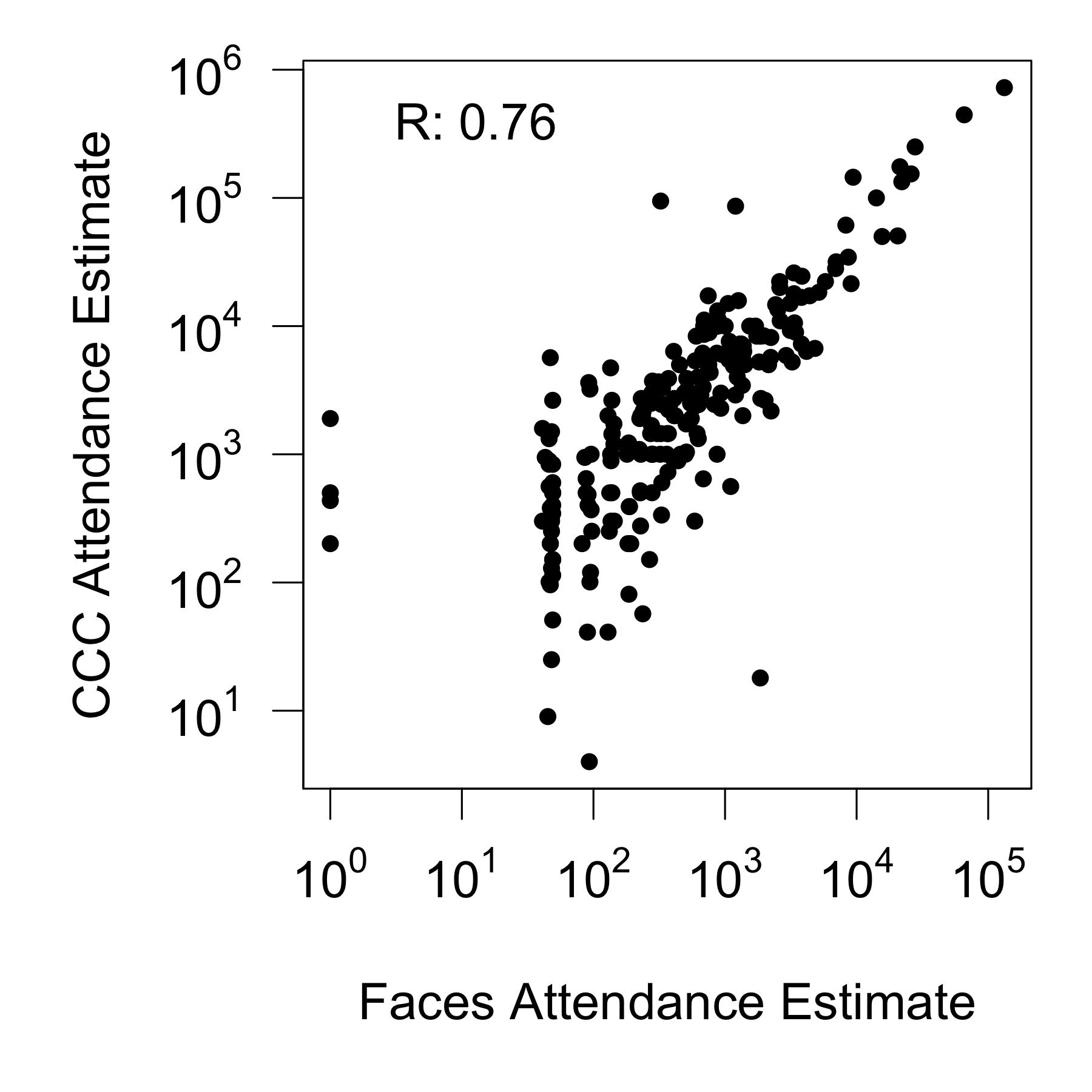} \\
(a) Logged Correlation
\end{figure}

\begin{figure}
\centering
\caption{Summing Number of Faces Accurately Measures Protest Size in South Korea}
\label{img:facesVerify_kr}
\includegraphics[width=.65\linewidth]{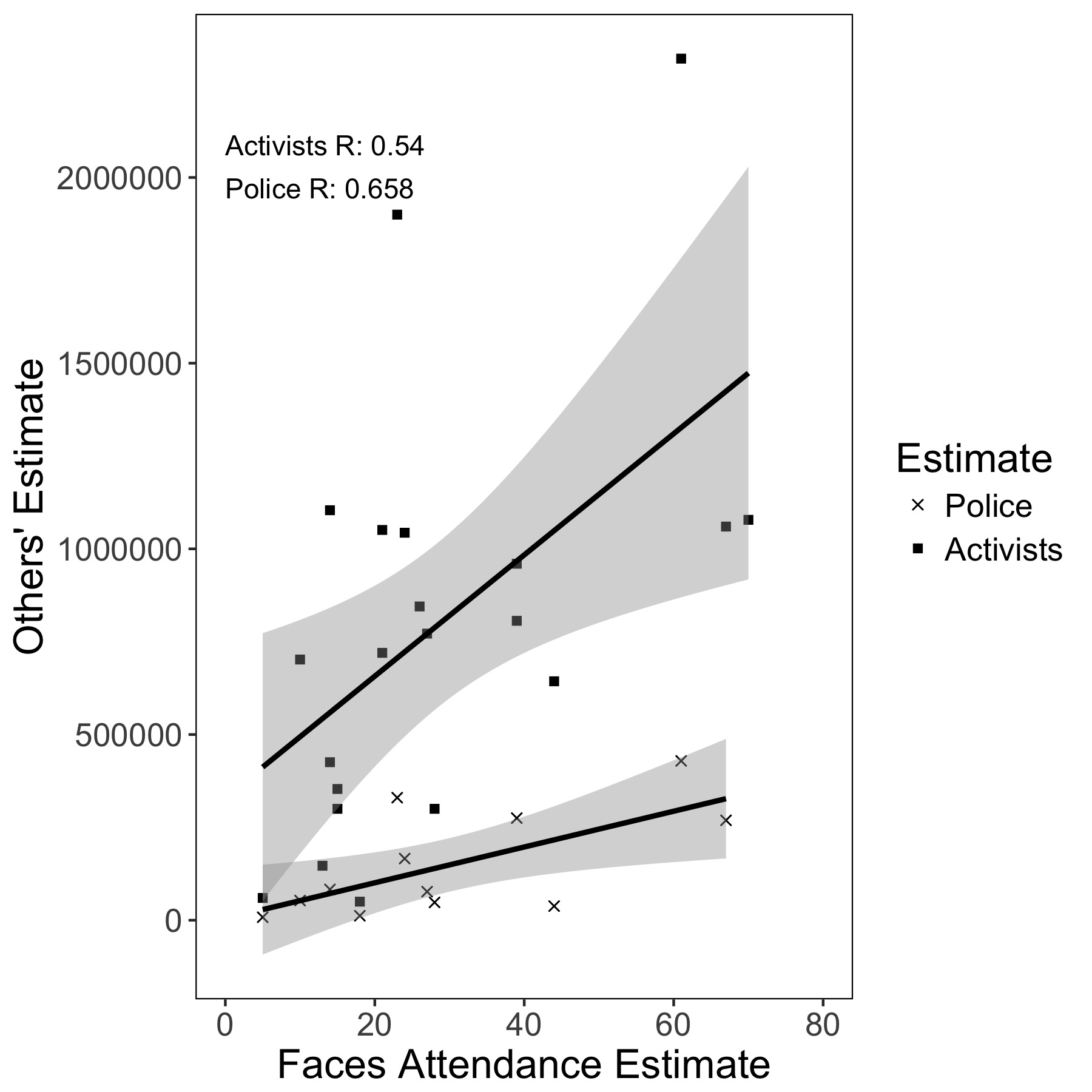} \\
(a) Raw Correlation \\
\includegraphics[width=.65\linewidth]{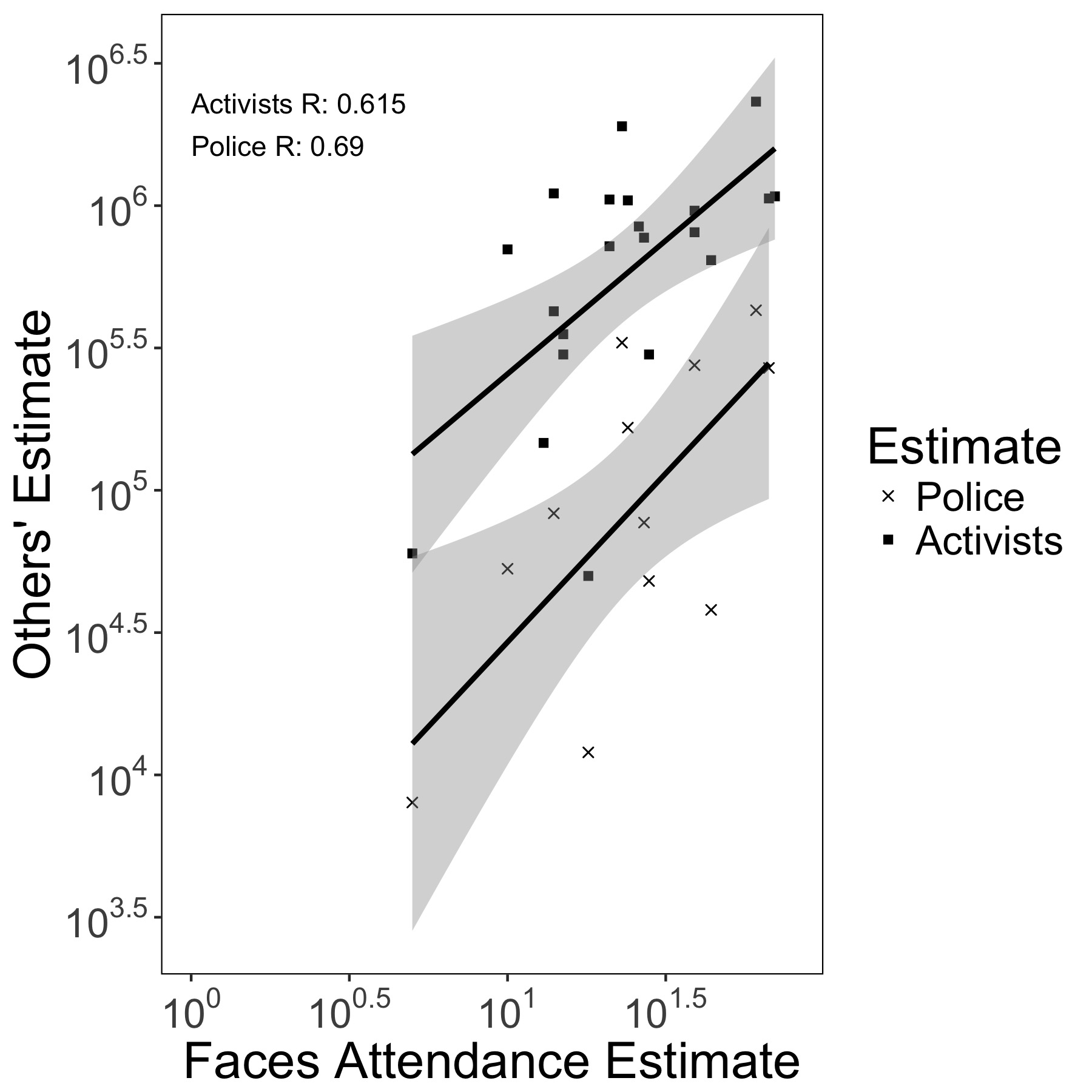} \\
(b) Logged Correlation
\end{figure}

We generate crowd size estimates by counting all faces in protest photos per day for a given location. Other studies have found that activity on Twitter correlates with verified estimates of crowd size for airports, stadiums, and protests \citep{Botta2015}.  Figure \ref{img:facesVerify} shows that our measure of protest size correlates with these protest estimates very well (.76 when logged) for the 2017 United States Women's March.  Figure \ref{img:facesVerify_kr} does the same for South Korea's 2016-2017 protests against President Park Geun-hye.  Using crowd size estimates provided by police and activists, as reported on Wikipedia, Figure \ref{img:facesVerify_kr} shows that the same procedure works in South Korea, though the police appear to provide more accurate estimates than activists.\footnote{The police provided fewer size estimates than activists, while this figure shows the correlation for all sizes each group reports.  Restricting the analysis to only those events for which police report size does not change the results.} 

Figurse \ref{img:facesVerify} and \ref{img:facesVerify_kr} suggest that computer vision can provide reliable estimates of protest size.

\clearpage

\subsection{South Korea and Hong Kong Protests in More Detail}
\label{skprotests}  

\begin{figure}[htbp!]
\caption{Change in Protest Size}
\centering
\begin{tabular}{c @{\qquad} c}
\includegraphics[width=.495\linewidth]{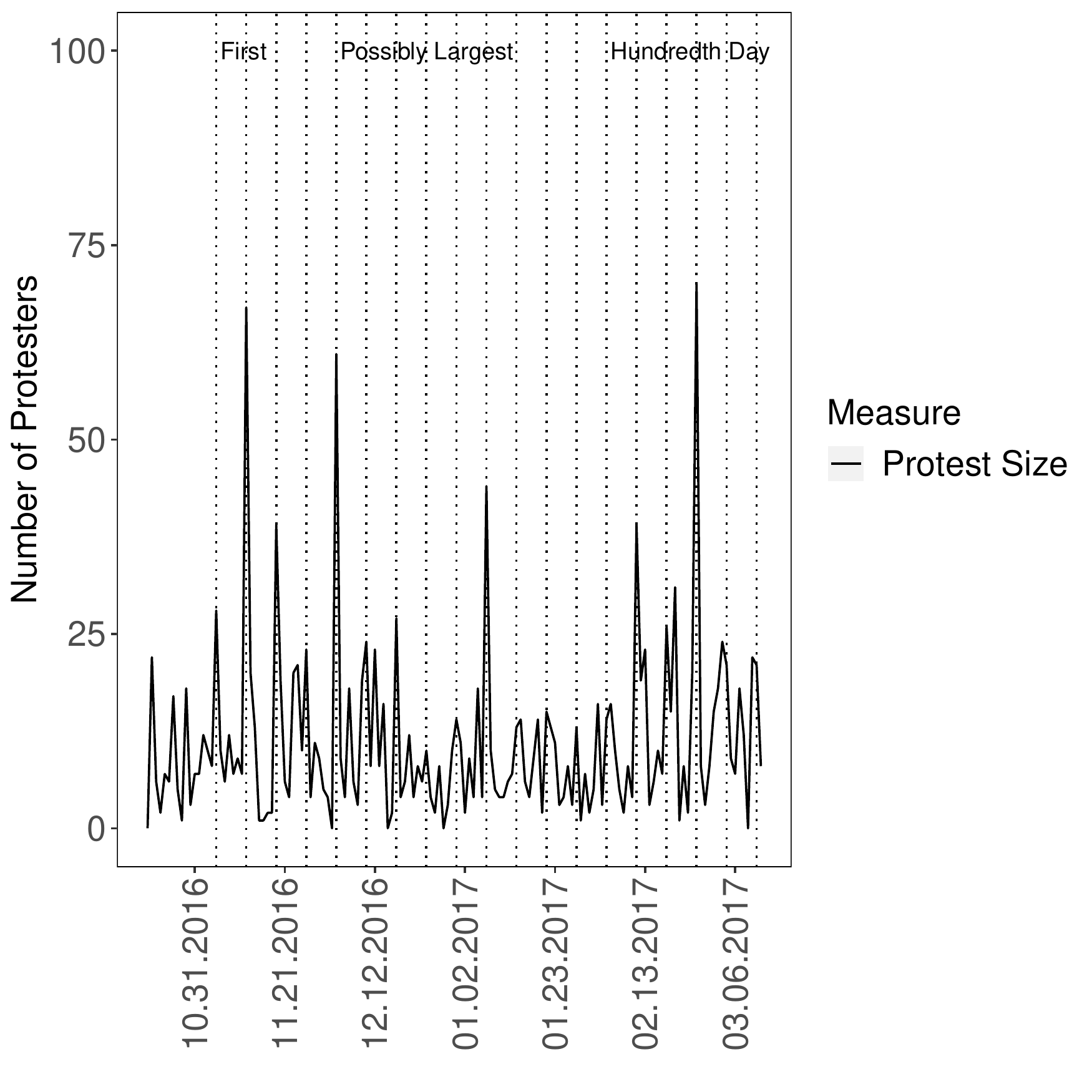} & 
\includegraphics[width=.495\linewidth]{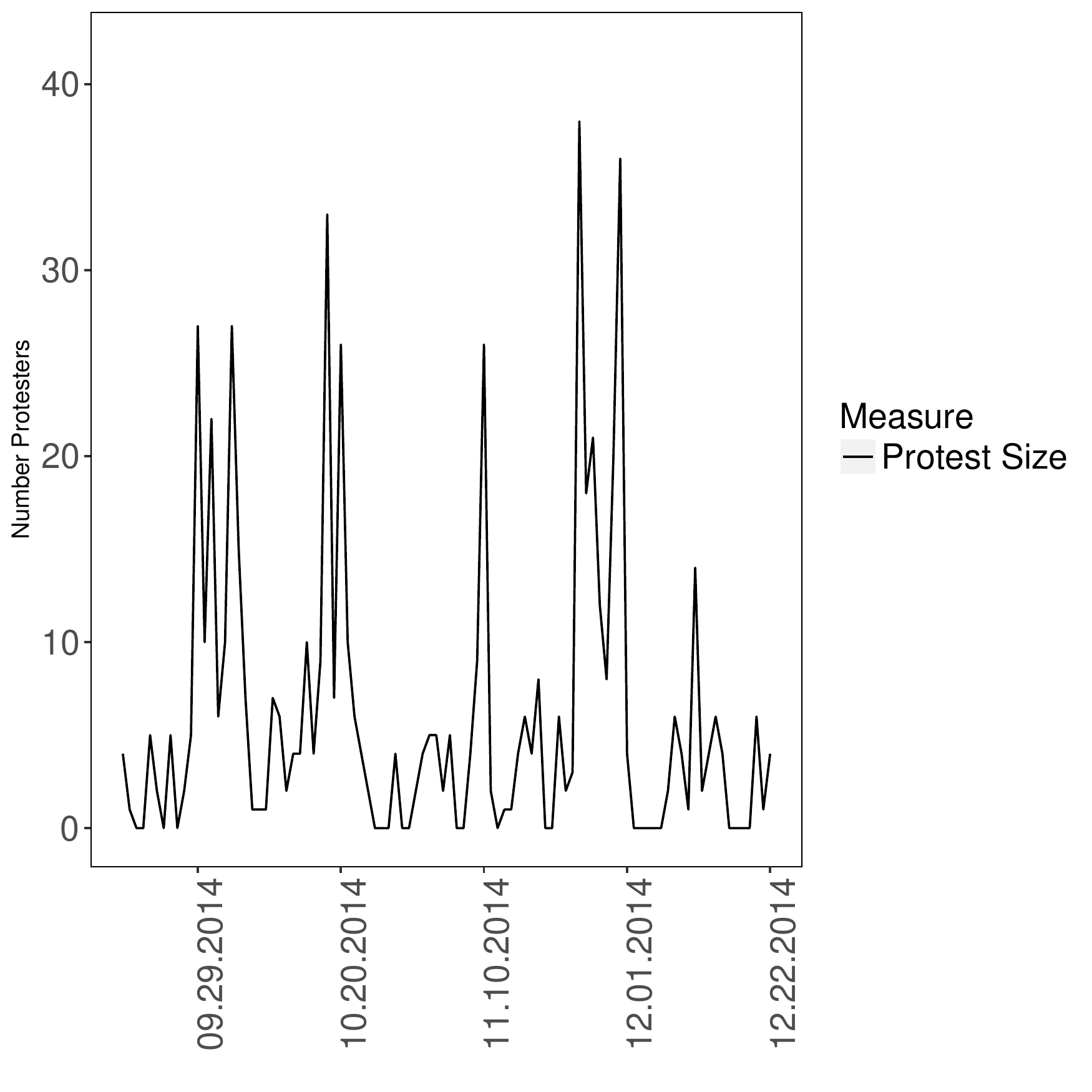} \\
(a) South Korea & (b) Hong Kong \\
\end{tabular}
\label{img:protestSize}
\end{figure}

Figure \ref{img:protestSize} shows daily level variation in the size of the South Korean and Hong Kong protests.  In South Korea, There are clear spikes on Saturdays (the dotted lines), and these spikes correspond to known protest events.  There is some discrepancy between the largest protest as reported by police and activists and as recorded via Twitter, but the overall patterns match others' estimates (see Figure \ref{img:facesVerify_kr}).

\begin{figure}[htbp!]
\caption{Percent of Female Faces}
\label{img:protestGender}
\centering
\begin{tabular}{c @{\qquad} c}
\includegraphics[width=.495\linewidth]{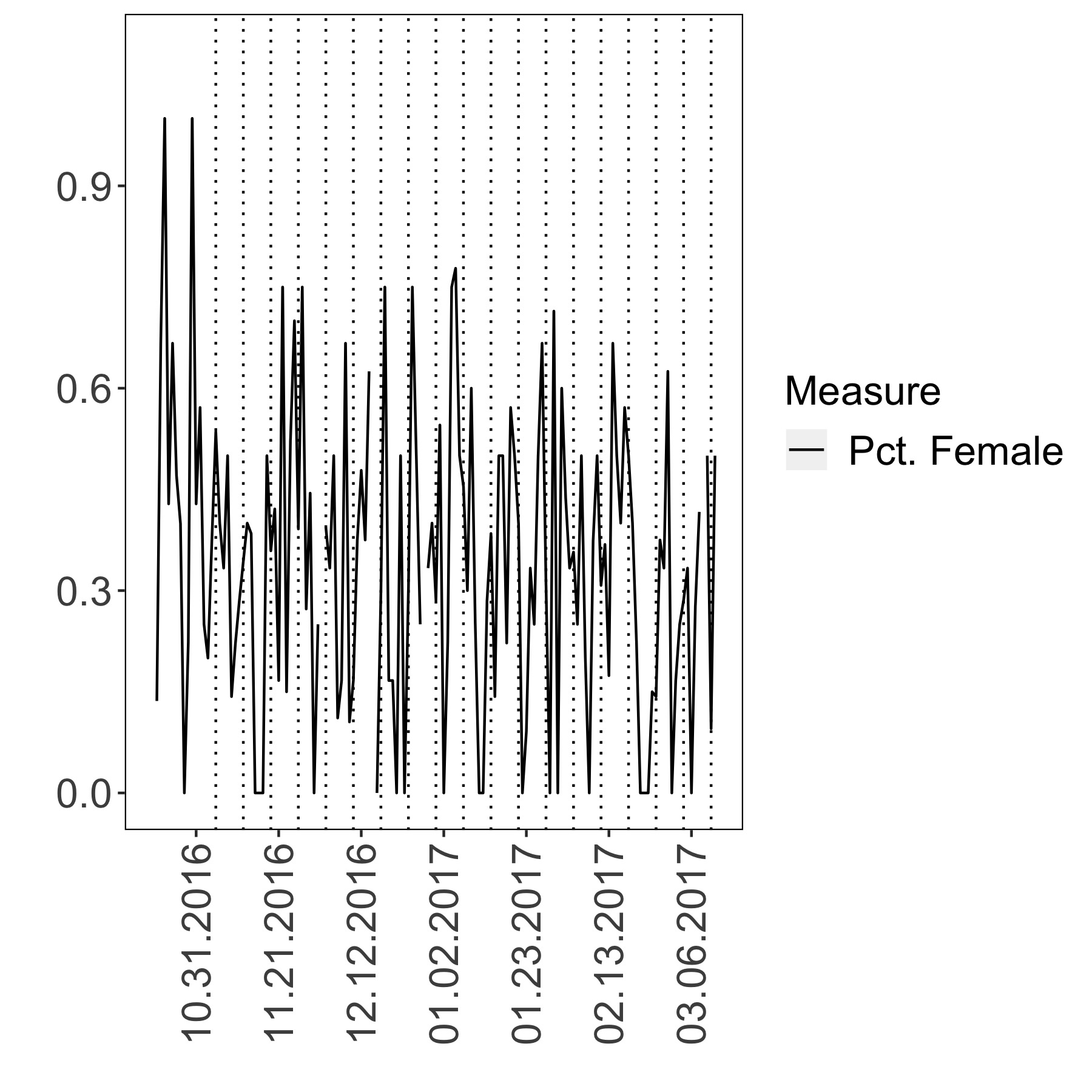} & 
\includegraphics[width=.495\linewidth]{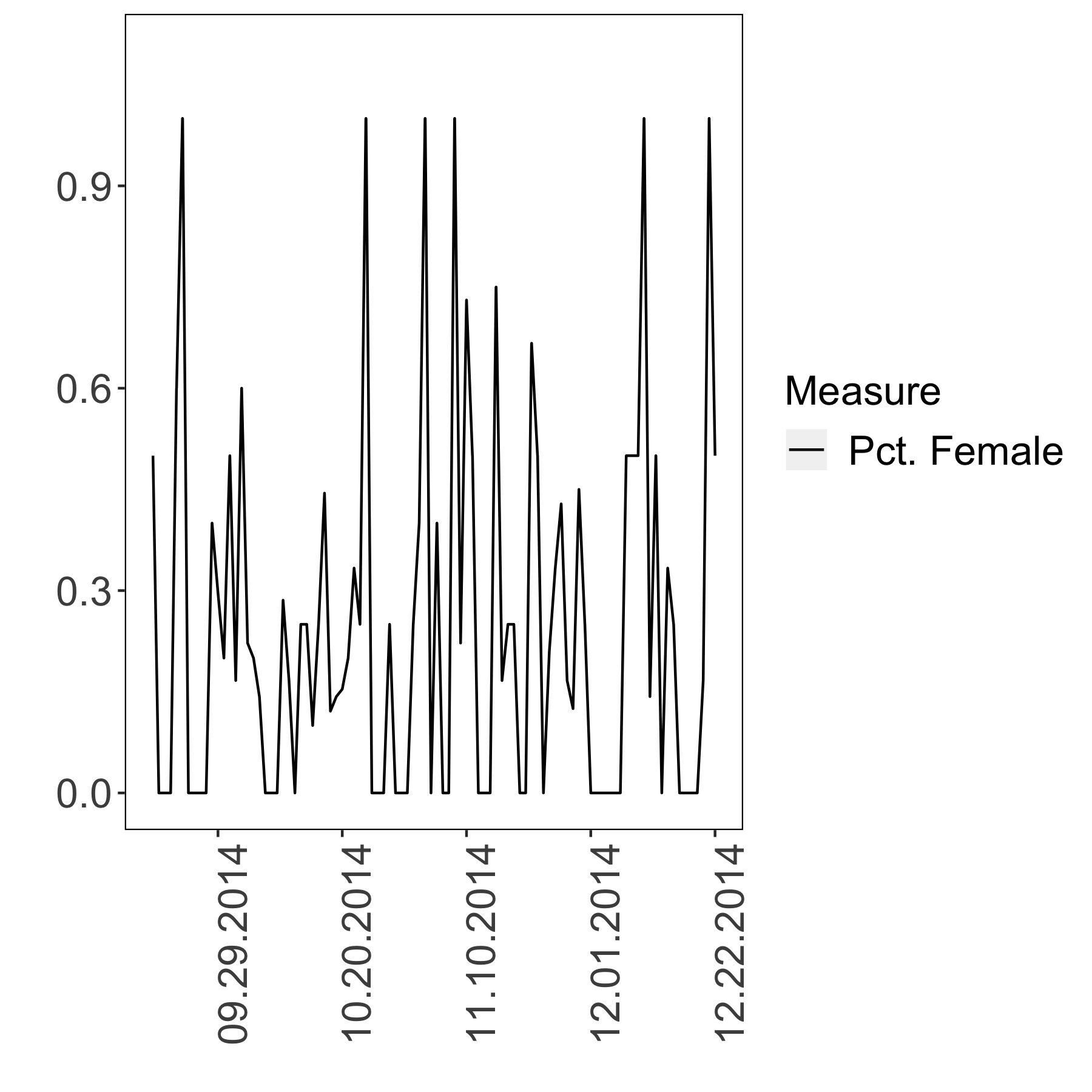}\\
(a) South Korea & (b) Hong Kong \\
\end{tabular}
\end{figure}

Figure \ref{img:protestGender} shows variation in the percent of protesters who are female at each protest.  The vertical dashed lines indicate Saturdays.

\begin{figure}[htbp!]
\caption{Percent of Photos with a Child}
\label{img:protestChild}
\centering
\begin{tabular}{c @{\qquad} c}
\includegraphics[width=.495\linewidth]{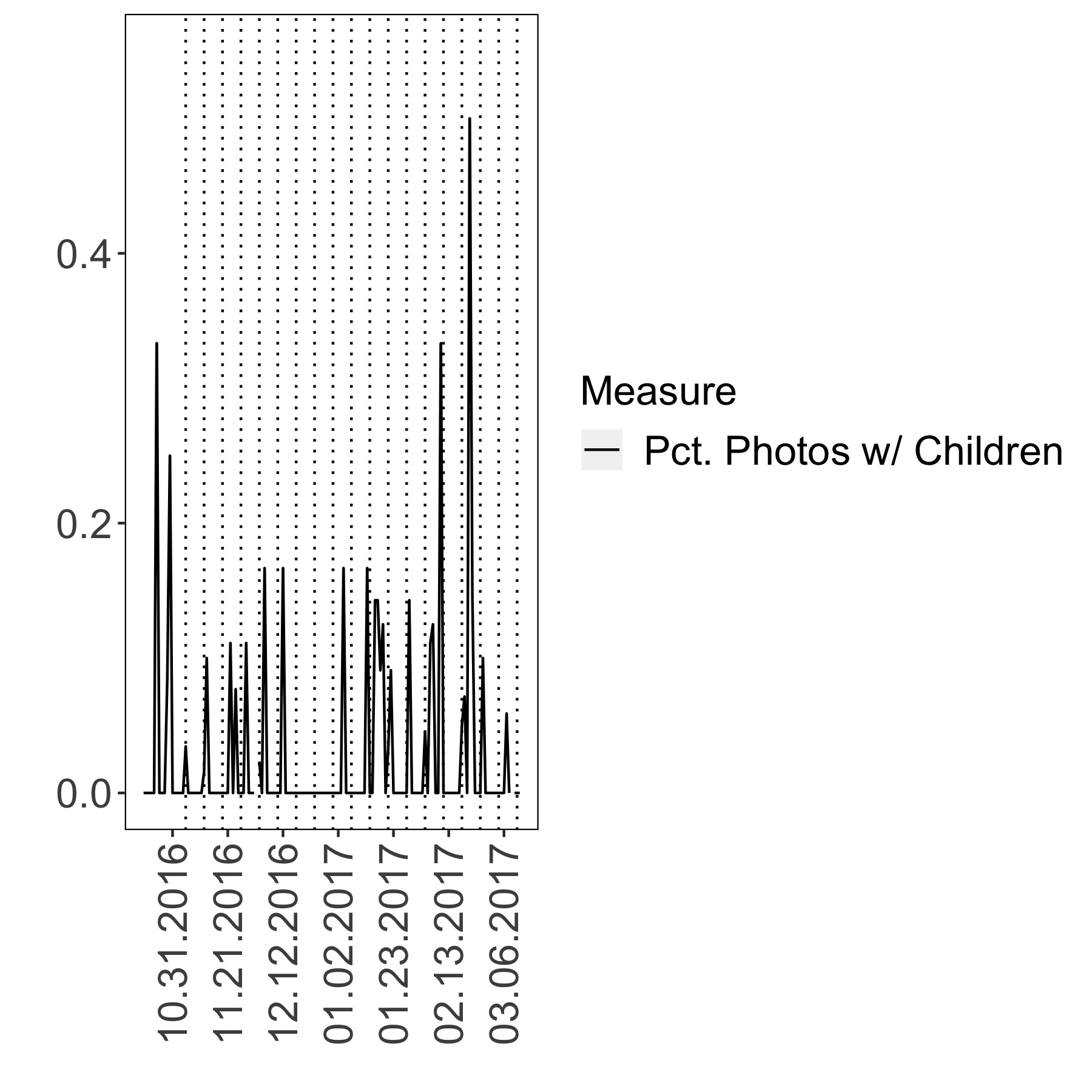} & 
\includegraphics[width=.495\linewidth]{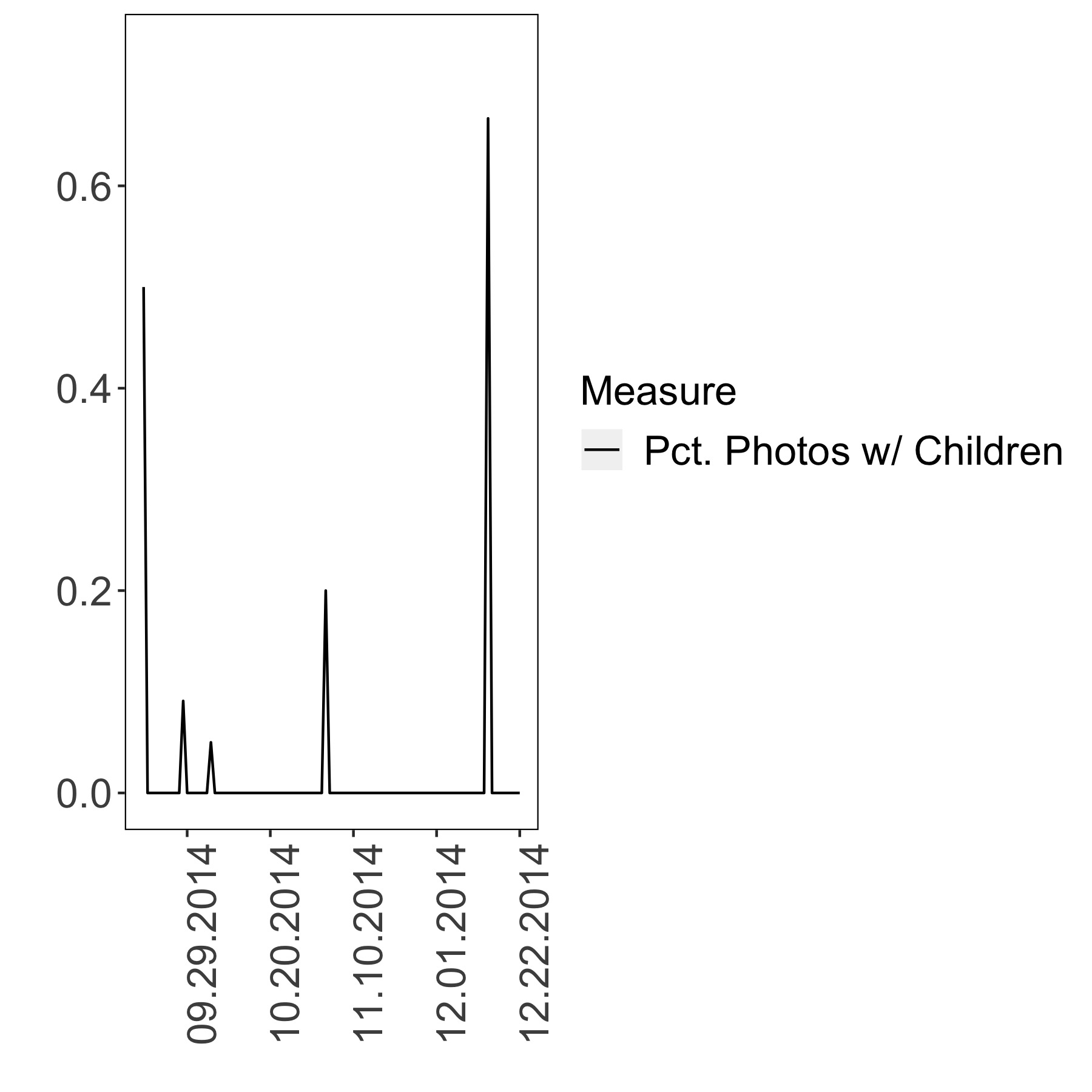}\\
(a) South Korea & (b) Hong Kong \\
\end{tabular}
\end{figure}

Figure \ref{img:protestChild} shows variation in the percent of images which contain a child.  Our classifier was not trained to recognize individual children, so we cannot count the number of children.  We are not aware of any event dataset that permits the study of protester demographics.

\begin{figure}[htbp!]
\caption{Perceived State Violence}
\label{img:protestStateViolence}
\centering
\begin{tabular}{c @{\qquad} c}
\includegraphics[width=.495\linewidth]{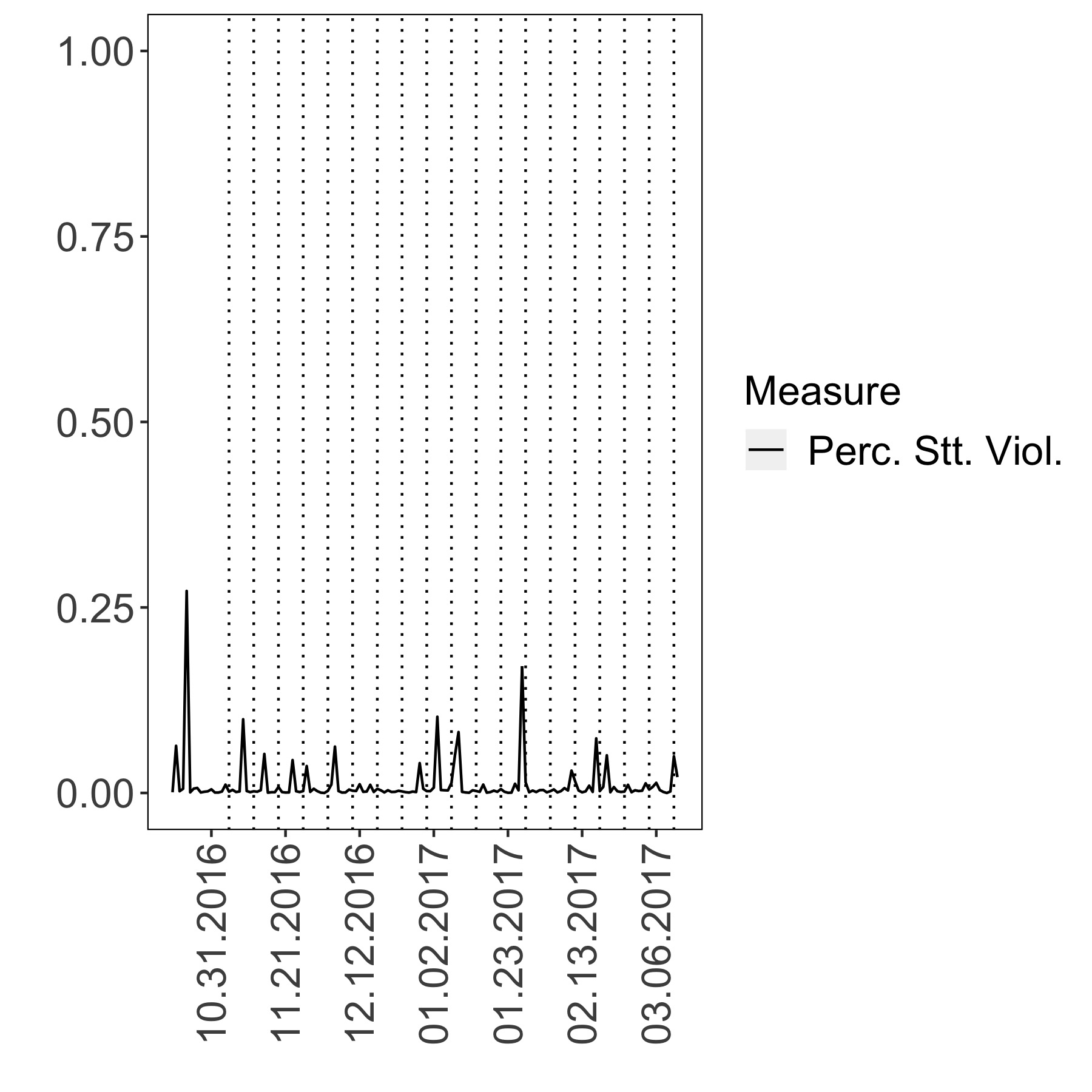} & 
\includegraphics[width=.495\linewidth]{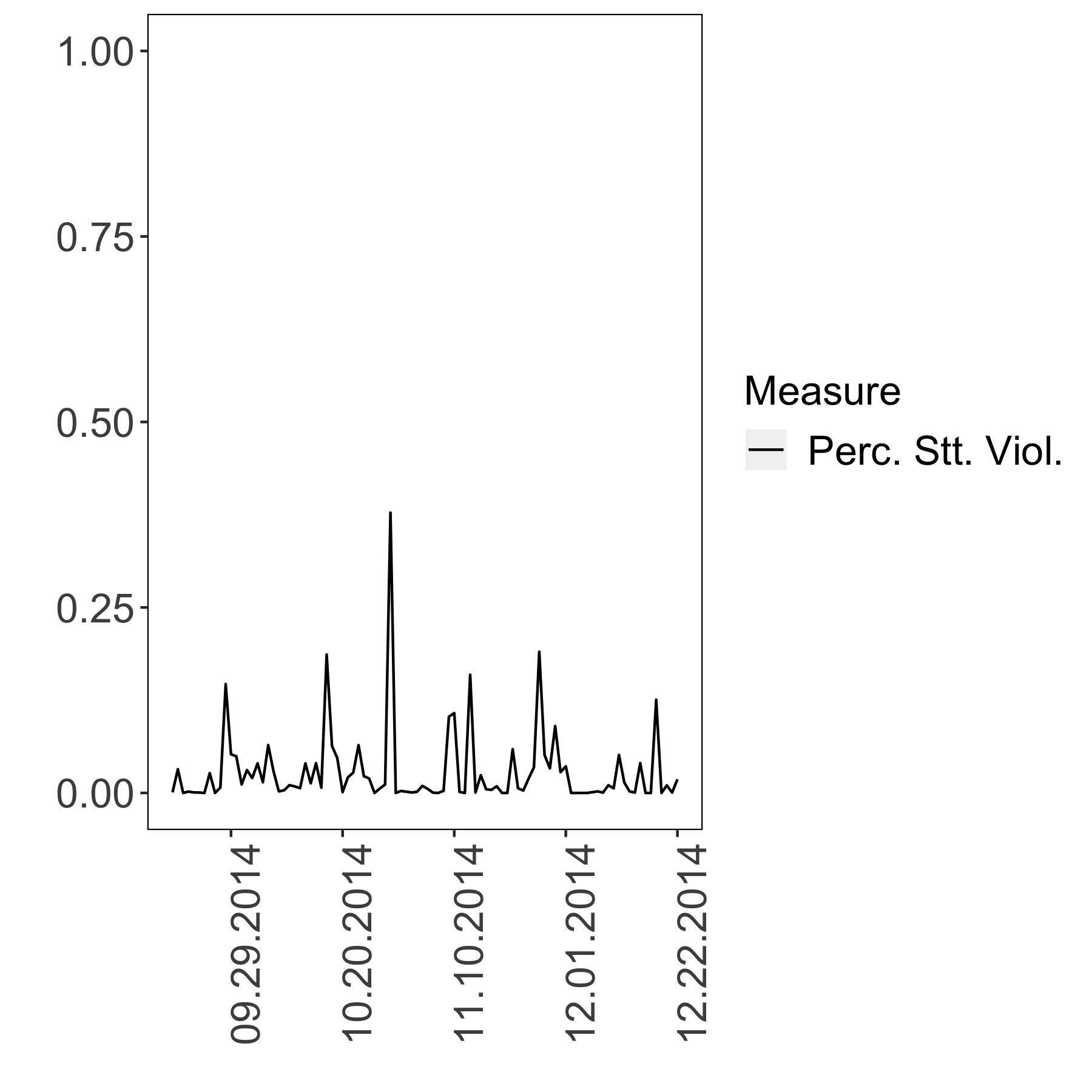}\\
(a) South Korea & (b) Hong Kong \\
\end{tabular}
\end{figure}

Figure \ref{img:protestStateViolence} shows the average perceived state violence during the two protests.  To generate these estimates, we presented pairs of 11,659 protest images to Amazon Mechanical Turk workers and asked each which image is more violent, creating 58,295 ratings ($\frac{11,659*10}{2}$).    From these ratings, the Bradley-Terry model generates a score from [0,1], where a higher rating equates with more violence \citep{bradley1952rank}.  Since violence is subjective, we call this outcome ``perceived violence".  An advantage of measuring violence from images, compared to the Goldstein Scale \citep{Goldstein1992}, is that it is continuously valued and does not require an image to document the interaction between two actors, as the Goldstein Scale does.

Figures \ref{img:protestSize}--\ref{img:protestStateViolence} suggest descriptive differences in the protest that match how the two protest waves were covered in newspapers.  The Korean protests were large and peaceful, whereas students dominated the Hong Kong protests, especially because police responded to protesters with tear gas and violence.  A t-test comparing perceived state violence during the two waves confirms that Hong Kong's protests had more state violence (.029 versus .007, $p=.0004$) and smaller crowds (6.24 versus 28.24, $p=.00003$).  Hong Kong protest photos also show fewer women and children in photos, though that result is not statistically significant.  These results hold when restricting the South Korea sample to only Saturdays.

Finally, we can use these data to model the correlations between various protest features and the size of protest.  To that end, we have modeled the size of the protests as a function of perceived violence (general, protester, and state), the presence of fire, the average number of faces in a photo, gender diversity, whether a protest image contains a child, the number of tweets with protest images, and the size of the most recent protest.  All variables are lagged by one day.  The dependent variable is the logarithm (base 10) of the number of faces in protest images; using the raw count does not change results.  All independent variables are standardized, so the coefficient represents the percent change in the dependent variable in response to a one unit change of the independent variable.  The data for Hong Kong and South Korea are pooled, though only South Korea's Saturdays are kept.  %For more details on the model, see \citet{Joo2018}.

\begin{figure}[htbp!]
\caption{Regression Results.  DV is $Log_{10}(Faces_{i})$.}
\label{img:regressionResults}
\centering
\includegraphics[width=.65\linewidth]{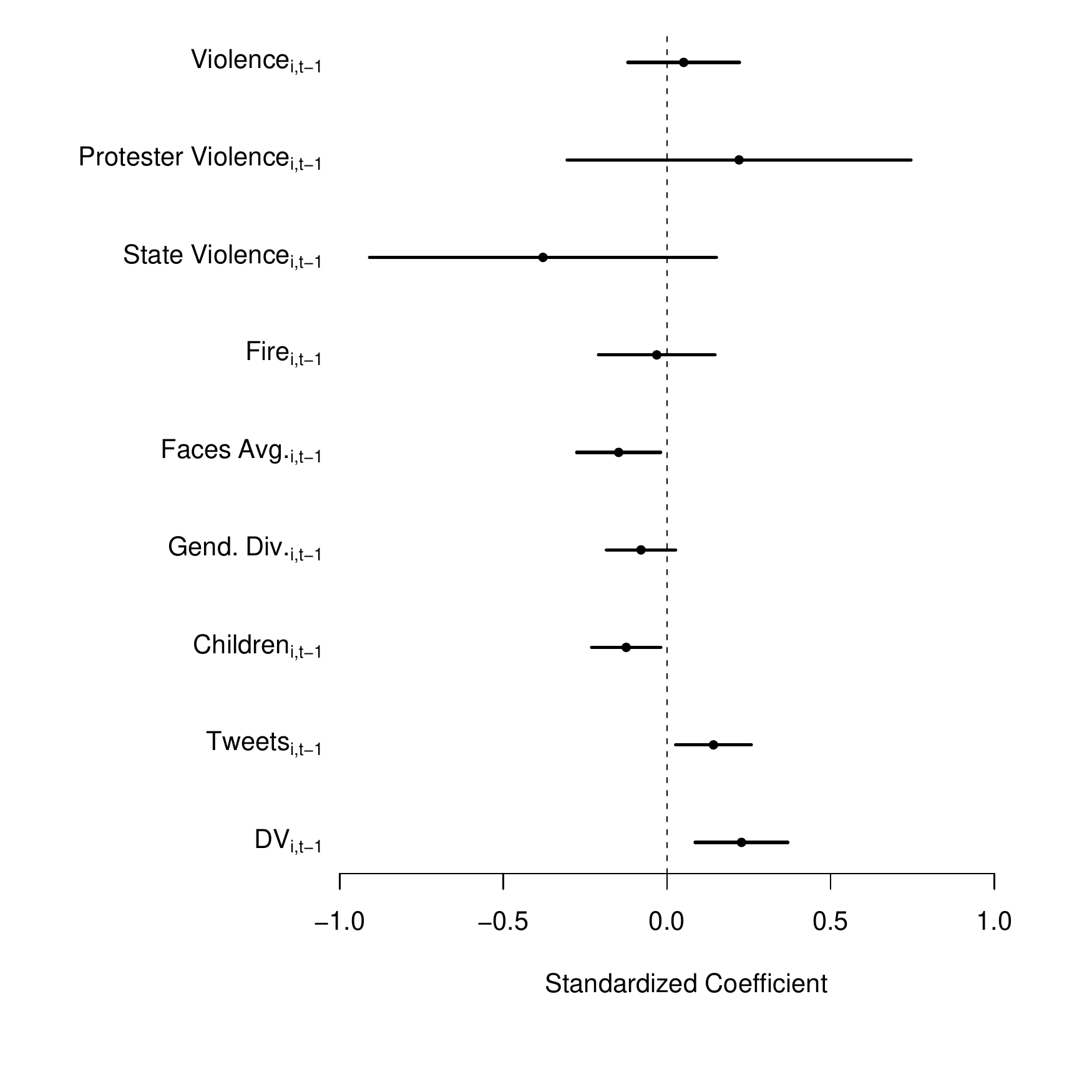}  \\
(a) South Korea and Hong Kong \\
\end{figure}

Figure \ref{img:regressionResults} shows the results of this model, with error bars representing 95\% confidence intervals.  The model finds autocorrelation of protest size.  More tweets with protest images also correlates with larger subsequent protests.  On the other hand, photos with more faces and of children correlate with smaller protests.  While the results for state violence and gender diversity do not reach conventional levels of statistical significance, they may with more observations.  It is not surprising that state violence could decrease protest size, but it is surprising that gender and age diversity would.  Though more investigation is necessary, we suspect that these variables are trailing indicators, that diversity increases as protests grow, not vice-versa.

\begin{figure}[htbp!]
\caption{Regression Results.  DV is $Log_{10}(Faces_{i})$.}
\label{img:regressionResults2}
\centering
\includegraphics[width=.65\linewidth]{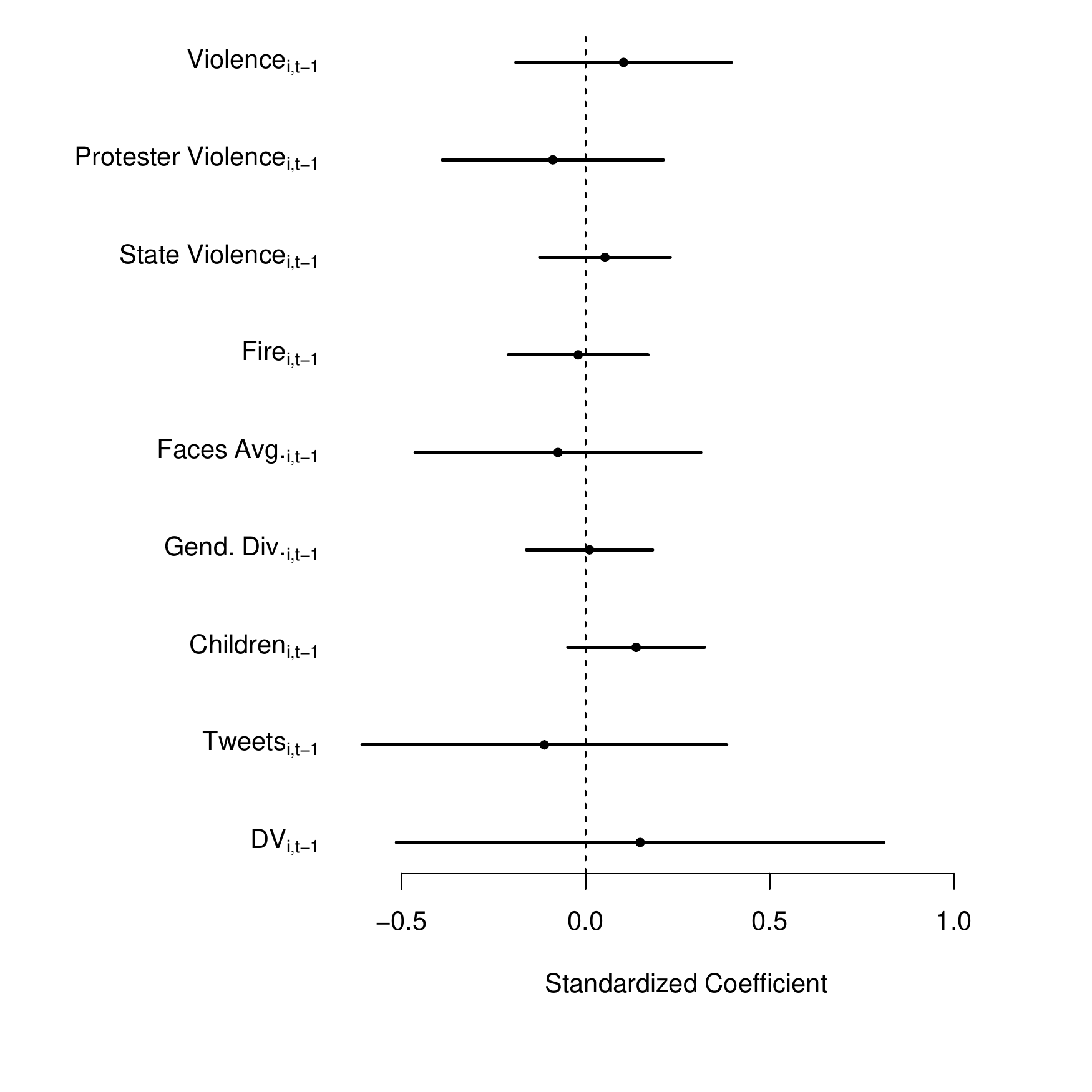}  \\
(a) South Korea \\
\includegraphics[width=.65\linewidth]{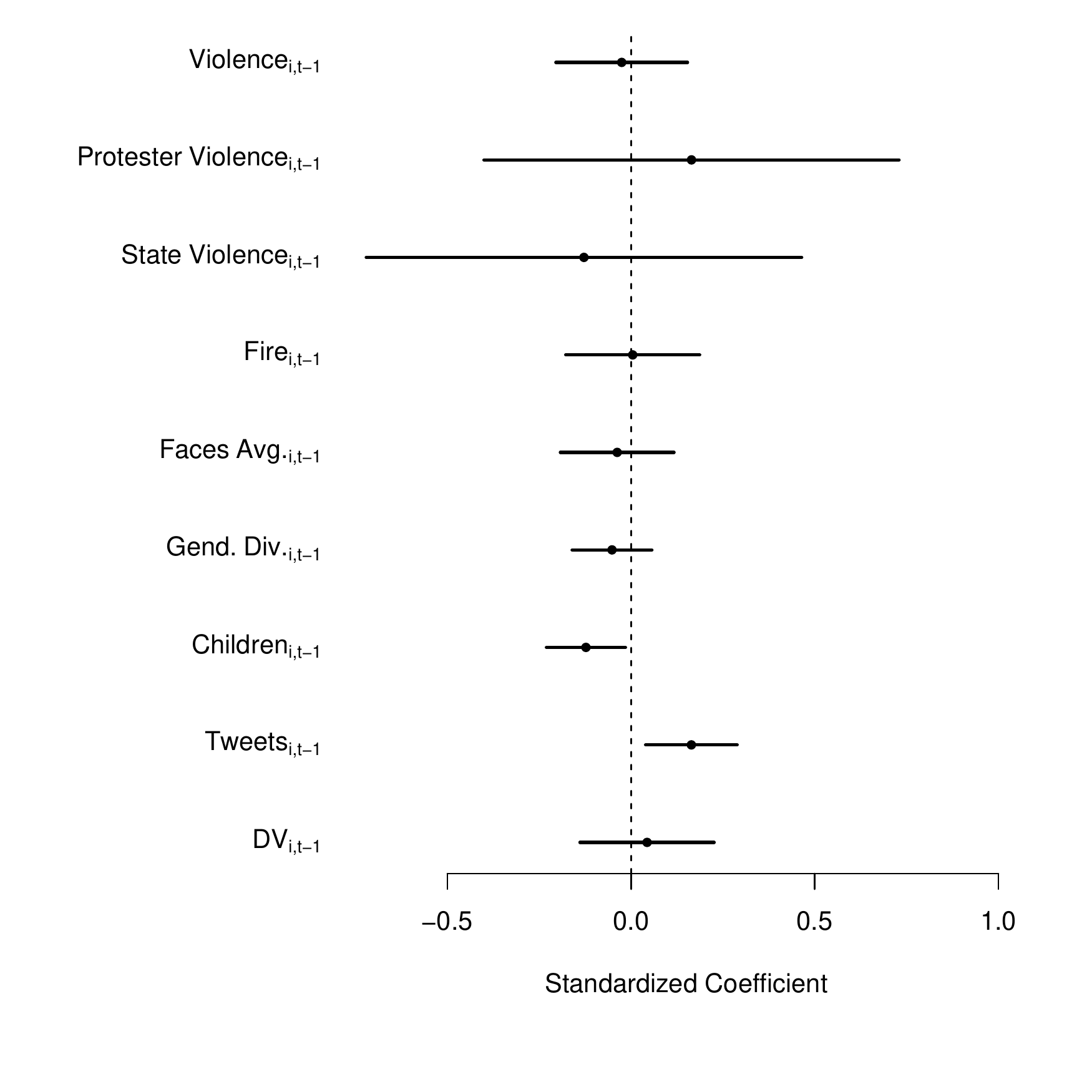}  \\
(b) Hong Kong \\
\end{figure}

Figure~\ref{img:regressionResults2} shows the results when modeling each country separately.  With fewer data, only two variables, Hong Kong's children and tweets, are statistically significant.

% \clearpage
\section{Conclusion}
% * <zst@luskin.ucla.edu> 2018-05-24T15:40:04.439Z:
% 
% NB: I use a version of the below paragraph in our violence paper.  I love it so much that I think it should be everywhere, and I have modified it enough that I think we are fine.
% 
% ^.
If a picture is worth 1,000 words, then it would require five kilobytes  of storage.   In fact, images from consumer cell phones and digital cameras require at least three megabytes of storage but usually more.  Even images shared on social media platforms, which are compressed from their original size, require hundreds of kilobytes of space.  A picture, in other words, is worth anywhere from 20,000 (100 kilobytes) to 5,000,000 words (5 megabytes).  In digital terms, it is more accurate to say that a picture is worth a book.
%\footnote{This estimate is poetic.  Another way to think of images is that they have high entropy, meaning they cannot be compressed as much as text.  The greater size of images reflects this greater difficulty of compressing them, not necessarily a ``true" quantum of information.} 

The extra information stored in images is both an opportunity and a challenge.  It is an opportunity because one image can document many more variables, including ones not measurable from text, than newspaper articles, speeches, or legislative documents.  The opportunity has remained underexploited because of the technical difficulty of identifying the objects and concepts encoded in an image, requiring researchers to rely on human coders.  Because human coders are slow, expensive, and have different interpretations of the same raw data (an image), studies using images have historically been small.  

Advances in machine learning algorithms, specifically the rise of convolutional neural networks,  have removed these barriers.  Along with increased hardware capabilities, especially the use of GPUs, these algorithms have expanded the frontier of computer capabilities.  For social media platforms, these advances mean automatically recognizing faces in uploaded images.  For governments, these advances mean increased biometric security as well as policing capabilities.  For researchers, these advances mean the ability to measure existing concepts better, operationalize measures previously only available in theoretical models, and do both with greater geographic and temporal resolution than previous efforts.

This paper has introduced computer vision and machine learning to political scientists.  It showed how convolutional neural networks process images, how to validate their classification output, and how these capabilities can contribute to various literatures.  There are certainly more applications for which space does not permit a discussion, and the applicability of these methods is limited largely by the imagination, and resources, of researchers.  As the third communication revolution continues to alter domestic and international politics \citep{Steele2002}, the ability to analyze the data produced by it will only grow in importance.

%Software libraries and online platforms to facilitate the integration of deep learning and computer vision into research are plentiful.  For both, PyTorch, TensorFlow, OpenCV, SimpleCV, and Theano are common Python libraries, while Keras is the primary R package.  Researchers can also submit images to online platforms such as Face++, the Google Vision API, Amazon Rekognition, and Open Face to receive annotation for images; these platforms are most developed for characterizing faces.  To classify non-standard attributes, such as whether an image is of a protest or contains police, the researcher will need to use software libraries and develop a custom neural network.  

Finally, the pipeline for generating data from images is the same as from text.  The researcher acquires a corpus of documents, sets aside some for training and testing, labels the training and test set into categories that fit the research question, and develops a model.  The model is then applied to the rest of the corpus, generating categories for each document, text or image, the researcher possesses.  These labels form the raw material of subsequent analysis.  In other words, research using text or images is much more similar than different.  Aside from hardware requirements, the main difference is the types of questions that can be answered.

\newpage
\singlespacing
\bibstyle{apsr}
\bibliography{main}

\end{document}